\documentclass[10pt,twocolumn,letterpaper]{article}

\usepackage{cvpr}              

\usepackage{microtype}

\renewcommand{\paragraph}[1]{\vspace{-0.37mm}\noindent\textbf{#1}}

\usepackage{mathtools}
\usepackage{parskip}

\usepackage{multirow}
\usepackage[table]{xcolor}
\definecolor{Prune}{gray}{0.45}

\usepackage{dsfont}

\newcommand{\real}{\mathbb{R}}

\newcounter{todocounter}

\definecolor{JoseBlue}{RGB}{0,90,180}

\usepackage{algpseudocode}
\usepackage{algorithm}

\usepackage{forloop}
\newcounter{ct}
\newcommand{\markdent}[1]{\forloop{ct}{0}{\value{ct} < #1}{\hspace{\algorithmicindent}}}
\newcommand{\markcomment}[2]{\Statex\markdent{#1} {\em #2}}

\newcommand{\N}{\mathbb{N}}
\newcommand{\vx}{\boldsymbol{x}}
\newcommand{\vlamb}{\boldsymbol{\lambda}}
\newcommand{\vrho}{\boldsymbol{\rho}}

\usepackage{siunitx}
\usepackage[hang,flushmargin]{footmisc}
\makeatletter
\newcommand\thefontsize[1]{{#1 The current font size is: \f@size pt\par}}
\makeatother

\definecolor{cvprblue}{rgb}{0.21,0.49,0.74}
\usepackage[pagebackref,breaklinks,colorlinks,allcolors=cvprblue]{hyperref}

\usepackage[table]{xcolor}
\usepackage{multirow}
\def\paperID{13262} 
\def\confName{CVPR}
\def\confYear{2026}

\newcommand{\ours}{CaCT\xspace}
\definecolor{darkgreen}{RGB}{0,120,0}

\newcommand{\misc}[1]{\textcolor{RoyalBlue}{#1}}

\title{Class Adaptive Conformal Training}

\author{%
Badr-Eddine Marani$^{1}$ \quad
Julio Silva-Rodr\'iguez$^{2}$ \quad
Ismail Ben Ayed$^{2}$ \\
Maria Vakalopoulou$^{1}$ \quad
Stergios Christodoulidis$^{1}$ \quad
Jose Dolz$^{2}$ \\[1ex]
$^{1}$ CentraleSup\'elec, Universit\'e Paris-Saclay \quad
$^{2}$ \'ETS Montr\'eal 
}

\usepackage{xr}
\begin{document}

\maketitle

\begin{abstract}
Deep neural networks have achieved remarkable success across a variety of tasks, yet they often suffer from unreliable probability estimates. As a result, they can be overconfident in their predictions. Conformal Prediction (CP) offers a principled framework for uncertainty quantification, yielding prediction sets with rigorous coverage guarantees. Existing conformal training methods optimize for overall set size, but shaping the prediction sets in a class-conditional manner is not straightforward and typically requires prior knowledge of the data distribution. In this work, we introduce Class Adaptive Conformal Training (\ours), which formulates conformal training as an augmented Lagrangian optimization problem that adaptively learns to shape prediction sets class-conditionally without making any distributional assumptions. Experiments on multiple benchmark datasets, including standard and long-tailed image recognition as well as text classification, demonstrate that \ours consistently outperforms prior conformal training 
methods, producing significantly smaller and more informative prediction sets while maintaining the desired coverage guarantees.
\end{abstract}

\section{Introduction}

Deploying machine learning based systems in high-stakes, real-world scenarios requires not only strong predictive performance but also reliable measures of uncertainty. Conformal Prediction (CP) has emerged as a principled approach to meet this requirement~\cite{vovk2003mcp,vovk2005book}. Given a pre-trained model, a calibration dataset, and a new test example $X_\mathrm{test} \in \mathcal{X}$ with unknown label $Y_\mathrm{test} \in \mathcal{Y}$,
{CP outputs prediction sets $\mathcal{C}(X_\mathrm{test}) \subseteq \mathcal{Y}$, which are agnostic to the model and data distribution, and}
statistically guaranteed to include the correct output with a user-defined mis-coverage level $\alpha \in [0, 1]$:
\begin{equation}
\label{eq:CP}
    \Pr(Y_\mathrm{test} \in \mathcal{C}(X_\mathrm{test})) \geq 1-\alpha.
\end{equation}
This distribution-free guarantee is a significant property, as it ensures the validity of the 
prediction sets.
Yet, while validity is crucial, 
the practical value of a prediction set depends on its efficiency, which is typically evaluated as the expected size of the predicted sets $\mathbb{E}_X |\mathcal{C}(X)|$. For instance, a trivial set that contains all labels is \textit{valid but uninformative}. Moreover, CP sets should exhibit \textit{adaptivity}, producing compact sets for instances deemed easy by the model and larger ones for more uncertain or ambiguous cases. 

\begin{figure}[t!]
    \centering
    \includegraphics[width=\linewidth]{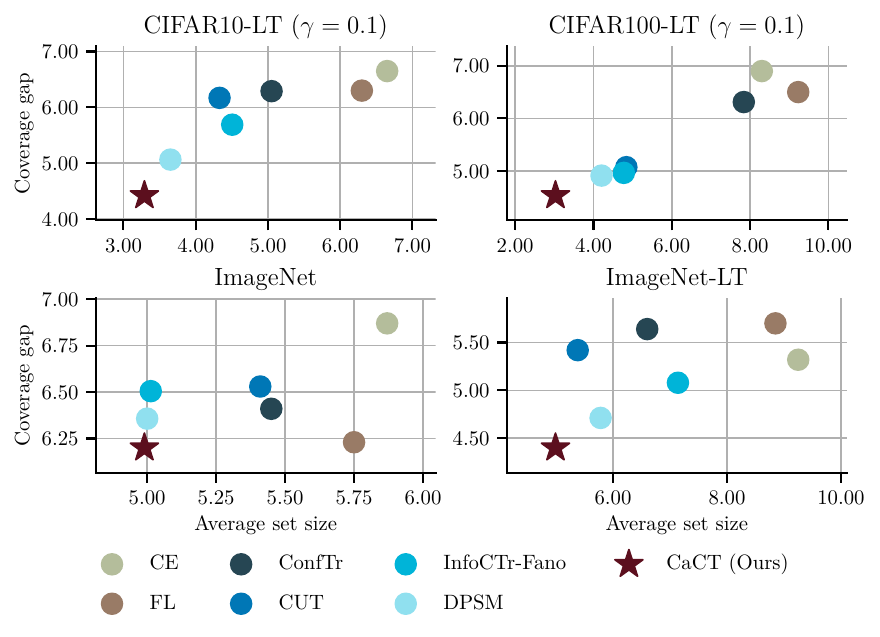}
    \caption{\textbf{Existing conformal training methods fail to account for datasets with a large number of classes or long-tailed distributions.} We compare \textbf{\ours} against conformal training methods, and report coverage gap and set size across relevant benchmarks. Lower values for both metrics indicate better conformalization and more informative prediction sets (\ie, better \textit{efficiency}) across all classes. Among all methods, \textbf{\ours} exhibits superior performance.}
    \label{fig:cov-size}
    \vspace{-3mm}
\end{figure}

Methods for enhancing the efficiency of conformal predictors span from designing effective non-conformity scores~\cite{angelopoulos2022raps,romano2020aps,sadinle2018thr} to explicit optimization-based strategies~\cite{stutz2022conftr,correia2025infocp,einbinder2022cut,bellotti2021conftr,shi2025direct}. While the former approach enjoys widespread adoption, 
applying CP solely as a post-training procedure prevents the underlying model from adapting to the construction of confidence sets~\cite{stutz2022conftr}. Consequently, training conformal models end-to-end has emerged as a particularly promising strategy for achieving both validity and improved efficiency.  
In particular, ConfTr~\cite{stutz2022conftr} represents one of the pioneer efforts to integrate model training with conformal prediction end-to-end. By resorting to smooth implementations of recent non-conformity scores, ConfTr
introduces a differentiable formulation of
CP, allowing the adoption of existing CP methods during training. Compared to standard training, this strategy reduces the average confidence set size of state-of-the-art CP approaches. 

\begin{figure}[t]
    \centering
    \includegraphics[width=\linewidth]{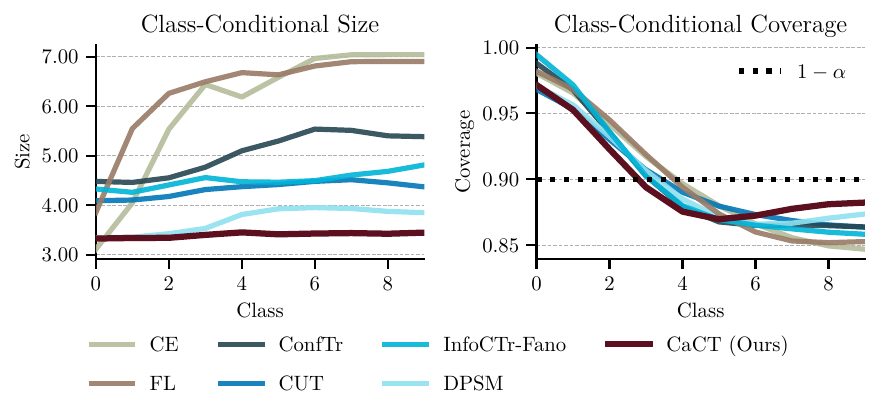}
    \caption{\textbf{Impact of per-class conformal set size and coverage with class frequency ($\alpha=0.1$) in CIFAR10-LT}. As classes become under-represented (toward the right of the x-axis), existing methods deteriorate their conformal performance, \ie, increase set sizes and decrease class-conditional coverage. In contrast, \ours is less sensitive to class imbalance, yielding smaller class set sizes (\textit{left}) while maintaining higher class coverage (\textit{right}) for these classes.}
    \label{fig:class-conditional-size-coverage}
    \vspace{-3mm}
\end{figure}

Nevertheless, although ConfTr~\cite{stutz2022conftr} advocates for improved efficiency, their adaptivity remains fundamentally constrained by the underlying formulation. 
To illustrate this limitation, let us examine its main learning objective, which introduces a regularization term to control efficiency by penalizing the size of the prediction set. This penalty is modulated by a single hyperparameter~$\lambda$, applied uniformly across all classes. While simple, 
this design makes the model’s performance highly sensitive to the choice of~$\lambda$, especially in large-scale or long-tailed settings. 
In addition, this weight is typically set before training and remains fixed, preventing the model from learning an optimal trade-off between accuracy and conformalization. Such a class-agnostic, rigid formulation is suboptimal, as different classes often exhibit distinct underlying characteristics. This limitation is further exacerbated in long-tailed scenarios, a common setting in real-world applications, where the majority classes often dominate the optimization dynamics, leading to biased conformalization and reduced adaptivity for underrepresented classes (\cref{fig:class-conditional-size-coverage}).

In light of the above limitations, we propose 
a novel conformal training strategy for \textit{efficient} and \textit{adaptive} conformal prediction. The key contributions can be summarized as: 
\begin{itemize}
    \item We introduce Class Adaptive Conformal Training (\ours), a novel framework that employs class-wise multipliers instead of a single global weight to construct \textit{efficient} and \textit{adaptive} prediction sets. This solves the issues highlighted above:
    \begin{enumerate*}[label=(\roman*)]
        \item scaling to datasets with a large number of classes, which exhibit diverse intrinsic difficulties, and
        \item effectively learning under class imbalance. 
    \end{enumerate*}
    \item The resulting constrained optimization problem is efficiently solved by the proposed modified Augmented Lagrangian algorithm, which provides adaptive and optimal weights for the class-wise penalties. 
    Compared to existing conformal training methods, 
    \ours shapes class-conditional prediction sets without designing new loss functions or requiring prior knowledge of the data distribution. \ours naturally handles datasets with many classes (\eg, ImageNet) and mitigates class imbalance issues (\eg, ImageNet-LT) by learning adaptive class weights.
    \item Comprehensive experiments across multiple benchmarks and distinct tasks, including standard image classification, long-tailed image recognition, and text classification, show that \ours produces smaller and more informative prediction sets, while maintaining 
    coverage, compared to existing training CP methods. As shown in~\cref{fig:cov-size}, \ours outperforms state-of-the-art conformal training methods when considering both set efficiency and coverage gap, particularly in more realistic large-scale datasets with a large number of classes or under class imbalance.
\end{itemize}

\section{Related Work}
\paragraph{Conformal Prediction (CP)} provides a principled framework to quantify uncertainty~\cite{angelopoulos2022intro, vovk2005book}.
Rather than returning a single label, CP outputs a prediction set that contains the true outcome with a user-specified probability.
For instance, in medical diagnosis, a conformal predictor might return a set of possible diseases and guarantee that the correct one is included 90\% of the time.
This framework has recently attracted significant attention~\cite{angelopoulos2022intro} and has been successfully applied across tasks including regression~\cite{romano2019cqr}, classification~\cite{angelopoulos2022raps, romano2020aps,silva2025confot,fillioux2025benchcpssm}, image segmentation~\cite{brunekreef2023kandinskycp, mossina2024confseg}, 
and language modeling~\cite{ye2024cpllmzeroshot}.
Split CP~\cite{papadopoulos2002inductive,vovk2005book} is a widely used method for constructing prediction sets.
Its core appeal lies in its simplicity and generality: given a pre-trained model and a calibration set, conformal prediction wraps around the model, outputting prediction sets with guaranteed coverage on a held-out sample under exchangeability, at a fraction of the computational cost, and without making any assumptions about the underlying data distribution.
Despite its broad applicability, CP is a post-hoc and model-agnostic procedure, and its effectiveness depends strongly on the quality of the underlying predictive model. When the model is not well aligned with the goals of CP, the resulting sets may be excessively large~\cite{stutz2022conftr}. Although such sets are technically valid, they are often uninformative~\cite{bellotti2021conftr}, which raises the question of how models could be trained to produce more informative and efficient prediction sets.

\paragraph{Conformal Training} incorporates regularization terms into the training objective in order to enforce conformal properties~\cite{einbinder2022cut, stutz2022conftr, correia2025infocp, bellotti2021conftr, shi2025direct}.
For example, SCPO-IC~\cite{bellotti2021conftr} and ConfTr~\cite{stutz2022conftr} simulate Split CP during training by introducing a regularization term that computes a differentiable approximation of the prediction set size, encouraging the model to construct smaller prediction sets. Direct Prediction Set Minimization (DPSM)~\cite{shi2025direct} further improves ConfTr by reformulating conformal training as a bi-level optimization problem. Instead of using a stochastic approximation for the quantile, it directly learns the quantile of the conformity scores.
Conformalized Uncertainty-aware Training (CUT)~\cite{einbinder2022cut} penalizes deviations from the uniform distribution of non-conformity scores. A major limitation of all of the aforementioned methods, however, is that the regularization weight is applied uniformly across all samples, regardless of class or difficulty, limiting their ability to adapt to intrinsic differences or imbalances among classes. Though these methods manage to reduce inefficiency marginally, they fail to reduce it in under-represented classes while maintaining coverage, as shown in~\cref{fig:class-conditional-size-coverage}.

\paragraph{Learning from long-tailed data} is a central challenge in machine learning, as standard models are biased toward frequent classes, leading to poor generalization on under-represented categories. 
In the context of CP, marginal coverage guarantees ensure that the true label lies in the prediction set on average across all classes. However, these guarantees do not protect minority classes, which can lead to systematically over- or under-covered predictions. Label-conditional CP~\cite{vovk2003mcp} mitigates this problem by constructing prediction sets separately for each class, effectively applying Split CP per-class. While this approach ensures the intended coverage, it often produces overly conservative and uninformative sets for rare classes, especially when the available data is limited~\cite{ding2023clustercp}.
Thus, existing conformal training methods largely overlook this challenge, as they optimize for marginal prediction set size rather than class-conditional efficiency.

\section{Background and Motivation}

\paragraph{Notations.}
For a positive integer $n$, let $[n] \coloneqq \{1, \dots, n\}$.
Let us consider a standard supervised multi-class classification setting, where data is drawn from a distribution $\mathcal{P}$ over $\mathcal{X} \times \mathcal{Y}$. Here, $\mathcal{X}$ denotes the input space, and $\mathcal{Y} = \{1, \dots, K\}$ is the set of class labels, with $K$ the total number of classes.
To keep the notation simple, we use $\mathcal{P}_X$ to denote the marginal distribution of $X$, and $\mathcal{P}_{X \mid Y=y}$ to denote the conditional distribution of $X$ given a label $y \in \mathcal{Y}$.
Note that we use uppercase $X,Y$ to denote input and output random variables, including samples in the training, validation, calibration, and testing sets, and fixed values with lowercase letters. This convention is maintained throughout the paper for consistency. So, let $\pi_\theta: \mathcal{X} \to \Delta_K$ be a soft classifier, \eg, softmax scores, with parameters $\theta$, where $\Delta_K$ denotes the $(K-1)$-dimensional probability simplex.
The training dataset is given by $\mathcal{D}_\mathrm{train}=\{(X_i, Y_i)\}_{i=1}^n$, while $\mathcal{D}_\mathrm{val}$, $\mathcal{D}_\mathrm{cal}$ and $\mathcal{D}_\mathrm{test}$ represent the validation, calibration and test datasets, respectively. We denote
the size of the calibration dataset as $m = |\mathcal{D}_\mathrm{cal}|$,
$\mathcal{I}_y=\{i \mid Y_i=y\}$ as the indices of samples corresponding to a class $y$, and $n_y=|\mathcal{I}_y|$.
 
\paragraph{Conformal Prediction (CP)} offers a principled framework for quantifying predictive uncertainty by constructing prediction sets that are guaranteed to contain the true label with high probability. Given a user-defined mis-coverage level $\alpha \in [0,1]$, a calibration dataset, a predictive model parameterized by $\theta$, and a new test pair $(X_\mathrm{test}, Y_\mathrm{test}) \sim \mathcal{P}$, CP ensures that the prediction set $\mathcal{C}_\theta: \mathcal{X} \to 2^{|\mathcal{Y}|}$ guaranteed to contain the true label with high probability at least $1-\alpha$, following~\Cref{eq:CP}.

The construction of the prediction set relies on a non-conformity score function $S: \mathcal{X} \times \mathcal{Y} \to \mathbb{R}$, which measures how atypical a candidate label $y$ is for a given input $x$ relative to previously observed data~\cite{vovk2005book}.
For the $i$-th sample, we denote the corresponding score as $S_i = S_\theta(X_i, Y_i)$. The prediction set is then constructed by thresholding the non-conformity scores with the empirical quantile $\widehat{q}_\theta$ computed from $\mathcal{D}_\mathrm{cal}$:
\begin{equation}
    \mathcal{C}_\theta(X_\mathrm{test}) = \left\{ y \in \mathcal{Y} \mid S_\theta(X_\mathrm{test}, y) \leq \widehat{q}_\theta  \right\},\nonumber
\end{equation}
where $\widehat{q}_\theta$ is the $\lceil (m+1)(1-\alpha) \rceil / m$-th order statistic of $\{S_{\theta,i}\}_{i=1}^m$.
Under the standard exchangeability assumption~\cite{vovk2005book}, this construction ensures that the prediction set covers the true label with probability at least $1-\alpha$.

The prediction set size of a given input $X$, also known as a set efficiency, is defined as:
\begin{equation}\label{eq:set-size-definition}
\begin{aligned}
    |\mathcal{C}_{\theta}(X)| = \sum_{y\in\mathcal{Y}} \mathds{1} \! \left[ y \in \mathcal{C}_\theta(X) \right]   = \sum_{y\in\mathcal{Y}} \mathds{1} \! \left[ S_\theta(X,y) \leq \widehat{q}_\theta \right].\nonumber
\end{aligned}
\end{equation}

\paragraph{Conformal Training.} From an optimization perspective, the goal is to learn a model that:
\begin{enumerate*}[label=(\roman*)]
\item makes accurate predictions, and
\item yields minimal prediction sets 
at test-time.
\end{enumerate*}
This leads to the following constrained optimization problem:
\begin{equation} \label{eq:conftr_constrained_problem}
    \begin{aligned}
        \min_\theta \quad & \mathcal{L}_\mathrm{cls}(\theta)\\
        \text{s.t.} \quad & |\mathcal{C}_{\theta}(X)| \leq \eta,
        \: \text{a.s. } X\sim\mathcal{P}_X,
    \end{aligned}
\end{equation}
where $\mathcal{L}_\mathrm{cls}(\theta)$ is a classification loss (\eg, cross-entropy) and $\eta$ is the target set size.
We can solve this problem using a penalty-based method, as in ConfTr~\cite{stutz2022conftr}, which converts 
~\eqref{eq:conftr_constrained_problem} into the following unconstrained optimization problem:
\begin{equation}\label{eq:conftr_unconstrained_relu}
    \min_\theta \quad \mathcal{L}_\mathrm{ConfTr}(\theta) \coloneqq
    \mathcal{L}_\mathrm{cls}(\theta) + \lambda \mathcal{L}_\mathrm{s}(\theta),
\end{equation}
in which $\lambda \in \mathbb{R}_+$ is a 
hyper-parameter that controls the importance of the second term, \ie, the size loss, defined as:
\begin{equation}
\label{eq:sizeloss}
\mathcal{L}_\mathrm{s}(\theta) = \mathbb{E}_X \left[\Omega\left( \mathcal{C}_{\theta}(X) \right)\right],
\end{equation}
with
\begin{equation}
    \Omega\left( \mathcal{C}_{\theta}(X) \right) = 
    \max\left( 0, |\mathcal{C}_{\theta}(X)| - \eta \right).
\end{equation}
The penalty term in~\eqref{eq:sizeloss} approximates prediction set size at a specific mis-coverage level $\alpha$ (\eg, 0.01) and penalizes sets larger than $\eta$.
Note that the objective of problem~\eqref{eq:conftr_unconstrained_relu} admits a straightforward unbiased estimator using
a training set of $n$
i.i.d. samples $\left\{\left(X_i, Y_i\right)\right\}_{i=1}^n$:
\begin{equation} \label{eq:conftr_unbiased_objective}
    \widehat{\mathcal{L}}_\mathrm{ConfTr}(\theta) =
    \widehat{\mathcal{L}}_\mathrm{cls}(\theta) +
    \lambda \,\frac{1}{n} \sum_{i = 1}^n \Omega\left( \mathcal{C}_{\theta}(X_i) \right)
\end{equation}

\section{Proposed Approach}

\subsection{Class-wise Constraints for Conformal Training}

The constraint in~\eqref{eq:conftr_constrained_problem} enforces an upper bound $\eta$ on the prediction set size {\em marginally} across all data points. However, when the number of classes is large, or under class imbalance, under-represented classes receive overly conservative prediction sets~\cite{fillioux2025benchcpssm} (~\cref{fig:class-conditional-size-coverage}). To address this issue, we first propose replacing the marginal constraint with class-conditional constraints, resulting in the following constrained problem:
\begin{equation} \label{eq:our_constrained_problem}
    \begin{aligned}
        \min_\theta \quad &\mathcal{L}_\mathrm{cls}(\theta)\\
        \text{s.t.} \quad & |\mathcal{C}_{\theta}(X)| \leq \eta,
    \: \text{a.s. } X \sim \mathcal{P}_{X \mid Y = k} \text{ for all } k \in \mathcal{Y},
    \end{aligned}
\end{equation}
which introduces $K = |\mathcal{Y}|$ constraints, one for each class, encouraging better control over class-wise efficiency. This modifies the learning objective of ConfTr~\cite{stutz2022conftr} with the proposed unconstrained problem defined as:
\begin{equation}
\label{eq:unconstrained_approx}
\min_\theta \quad
\widehat{\mathcal{L}}_\mathrm{cls}(\theta) +
\lambda \,\frac{1}{n} \sum_{y\in\mathcal{Y}}\sum_{i\in\mathcal{I}_y} \Omega(\mathcal{C}_{\theta}(X_i)),
\end{equation}
where $\mathcal{I}_y = \{i \mid Y_i = y\}$ denotes the set of indices for samples belonging to class $y \in \mathcal{Y}$. Note that in this case, the size constraint is enforced over each category in $\mathcal{I}_y$, and not marginally over all classes as in ConfTr.

Even though the above constraints account for set sizes in a class-wise fashion, treating all constraint violations equally with a single uniform penalty weight $\lambda$ fails to accurately solve the associated constrained problem in~\eqref{eq:our_constrained_problem}. Indeed, it is approximated by the unconstrained problem in~\eqref{eq:unconstrained_approx}, which uses a single uniform penalty weight, regardless of the category\footnote{An analogous argument holds for ConfTr~\cite{stutz2022conftr}: its constrained formulation in~\eqref{eq:conftr_constrained_problem} is only approximated by the unconstrained objective in~\eqref{eq:conftr_unbiased_objective}.}. This 
implicitly assumes that all classes have similar characteristics. However, in practice, classes exhibit vastly different behaviors: some classes may naturally produce smaller prediction sets due to high model confidence, while others, particularly minority classes in long-tailed distributions, may consistently violate the size constraint. Therefore, an enhanced training strategy would involve considering distinct penalty weights, one per category. This 
requires integrating 
$K$ penalty weights, 
resulting in the following learning objective:
\begin{equation}
\label{eq:classwise_constraints}
\min_\theta \quad
\widehat{\mathcal{L}}_\mathrm{cls}(\theta) +
\frac{1}{n} \sum_{y\in\mathcal{Y}} \lambda_y \sum_{i\in\mathcal{I}_y} \Omega(\mathcal{C}_{\theta}(X_i)),
\end{equation}
where $(\lambda_y)_{1\leq y \leq K} \in \real^K_+$, and each 
weight $\lambda_y$ balances the contribution of its associated class-wise penalty for category $y$ in the main learning objective. While this formulation 
allows controlling set sizes per-class, determining appropriate values for the $K$ penalty weights is not straightforward. Traditional 
grid search over $\boldsymbol{\lambda}$ quickly becomes infeasible, \eg, 
with $K = 1000$ classes (as in ImageNet~\cite{russakovsky2015imagenet}), 
even 5 values per dimension require evaluating $5^{1000}$ configurations. This is further exacerbated under long-tailed distributions, as different classes need penalty weights on different scales. In the following, we present 
an augmented Lagrangian strategy that automatically learns the class-wise penalty weights.

\subsection{Class Adaptive Conformal Training}
\label{ssec:our_method}

\paragraph{General Augmented Lagrangian Methods (ALM).}
To automatically learn penalty weights, ALM --which are widely used in the broader field of optimization~\cite{Bertsekas1996,NocedalWright2006}-- offer an appealing approach, integrating penalty functions and primal-dual updates to solve constrained optimization problems.

We begin by considering a generic constrained optimization problem ($\vx \in \real^d$ denotes a generic optimization variable in the Augmented Lagrangian formulation):
\begin{equation}\label{eq:alm_const_generic_prob}
    \min_{\vx} \quad
    f(\vx) \quad \text{subject to} \quad h_i(\vx) \leq 0, \quad i \in [m],
\end{equation}
where $f: \mathbb{R}^d \to \mathbb{R}$ is the \textit{objective function}, and $h_i: \mathbb{R}^d \to \mathbb{R}$ for $i = 1, \dots, m$ represent the \textit{set of constraint functions}.
One classical strategy is to convert Problem~\eqref{eq:alm_const_generic_prob} into a sequence of $j \in \mathbb{N}$ unconstrained problems, where each problem is solved approximately w.r.t $\vx$:
\begin{equation}\label{eq:alm_unconst_pen_generic_prob}
\begin{aligned}
    \min_{\vx} \quad \mathcal{L}^{(j)}(\vx)=
    f(x) 
    + \sum_{i=1}^m P(h_i(\vx), \lambda_i^{(j)}, \rho_i^{(j)}),    
\end{aligned}
\end{equation}%
where $P: \mathbb{R} \times \mathbb{R}_{++} \times \mathbb{R}_{++} \to \mathbb{R}$ is a \textit{penalty-Lagrangian function} $P(z, \lambda, \rho)$\footnote{We provide in~\Cref{sec:penalties_axioms} the axioms that must be satisfied to consider a penalty function~\cite{Birgin2005}.}, whose partial derivative with respect to its first argument exists, is positive, and continuous for all \( z \in \mathbb{R} \) and \( (\rho, \lambda) \in (\mathbb{R}_{++})^2 \). 
The parameters $\boldsymbol{\lambda}^{(j)}=(\lambda_i^{(j)})_{1\leq i \leq m} \in \mathbb{R}_{++}$ and $\boldsymbol{\rho}^{(j)}=(\rho^{(j)}_i)_{1\leq i \leq m} \in \mathbb{R}_{++}$ are the Lagrange multipliers and penalty coefficients associated to $P$ at iteration $j$, respectively. Minimizing $\mathcal{L}^{(j)}$ is referred to as the \textit{inner} iterations, whereas the \textit{outer} iterations denote the sequence of unconstrained problems. 

After solving Problem~\eqref{eq:alm_unconst_pen_generic_prob}, the penalty multipliers $\boldsymbol{\lambda}^{(j)}$ are updated by differentiating the penalty function with respect to its first argument (\ie, last inner minimization):
\begin{equation}\label{eq:alm_update_multipliers}
    \lambda_i^{(j+1)} = P^\prime(h_i(\vx^{(j)}), \lambda_i^{(j)}, \rho_i^{(j)}).
\end{equation}
This update rule corresponds to a first-order multiplier estimate for the constrained problem, where $\vx^{(j)}$ is the solution obtained at iteration $j$. To adapt the penalty parameters $\boldsymbol{\rho}^{(i)}$ during training, the behavior of the constraint function is monitored. If the constraint does not show improvement between iterations, the corresponding penalty parameter is scaled up by a factor $(\beta>1)$. This dynamic adjustment encourages the model to better satisfy constraints over time. 

In convex settings, alternating between approximate minimization of (\ref{eq:alm_unconst_pen_generic_prob}) and updates to the multipliers (\ref{eq:alm_update_multipliers}) leads to convergence toward a feasible solution. The inner loop targets the primal formulation, while the outer loop progressively refines the dual variables. Although theoretical convergence guarantees for ALM are primarily established in convex optimization, these techniques have also demonstrated strong empirical performance in nonconvex scenarios~\cite{Birgin2005}. Despite their potential, ALM-based approaches remain relatively overlooked in the deep learning context~\cite{rony2021augmented,liu2023cals,sangalli2022alm,silva2024closer}, and more so in the CP 
literature. This gap presents an opportunity to explore their effectiveness 
where standard strategies may struggle.

\paragraph{Integrating ALM Into Conformal Training.} We propose in this work to leverage ALM to solve the unconstrained problem in (\ref{eq:classwise_constraints}). Specifically, we reformulate Eq. (\ref{eq:classwise_constraints}) by integrating a penalty function $P$ parameterized by $(\boldsymbol{\rho},\boldsymbol{\lambda})\in \real^K_{++} \times \real^K_{++}$, which can be formally defined as:
\begin{equation}
    \min_{\theta, \boldsymbol{\lambda}} \quad \mathcal{L}_\mathrm{cls}(\theta) + \sum_{k=1}^K P\left(\widehat{d}_k - \eta, \lambda_k, \rho_k \right),
\end{equation}
where $\widehat{d}_k = \frac{1}{n_k}\sum_{i\in\mathcal{I}_k} |\mathcal{C}_{\theta}(X_i)|$ is the average class-wise prediction set size. 
Directly estimating Lagrange multipliers from the training set could easily lead to overfitting the training data.
We propose to leverage a validation set $\mathcal{D}_\mathrm{val}$ to reliably estimate the penalty multipliers at each epoch. In particular, each training epoch approximates the minimization of the primary loss function, while the average penalty multiplier value is subsequently estimated as an expectation over the calibration set. We formally define this procedure, where after training epoch $j$ the penalty multipliers for epoch $j+1$ are:
\begin{equation}
    \label{eq:lambda_update}
    \lambda_k^{(j+1)}=
    P' (\widehat{d}_k-\eta, \lambda_k^{(j)},\rho_k^{(j)})
\end{equation}

In this training scheme, the penalty multiplier is adjusted dynamically, increasing when the constraint is violated and decreasing when the constraint is satisfied. This mechanism effectively provides an adaptive strategy for selecting the penalty weight in a penalty-based method. 

Furthermore, we update penalty parameters $\vrho$ as:
\begin{equation} \label{eq:rho_update_rule}
    \rho_{k}^{(j+1)} =
    \begin{cases}
        \beta \rho_{k}^{(j)} \quad & \text{if} \quad \widehat{d}_k^{(j)} - \eta> \max {(0, \widehat{d}_k^{(j-1)}-\eta)}\\
        \rho_{k}^{(j)} \quad & \text{otherwise},
    \end{cases}
\end{equation}
where we compute the average constraint value per-class on the validation set $\mathcal{D}_\mathrm{val}$. Thus, for each category, if the average constraint remains positive and has not decreased compared to the previous epoch, the corresponding penalty parameter is multiplied by a scaling factor $\beta$. This update strategy ensures that $\vrho$ grows when constraint violations persist and stabilizes otherwise, effectively implementing an adaptive penalty adjustment process. 

Last, the choice of an appropriate penalty function is paramount to the performance of ALM methods, particularly in non-convex scenarios~\cite{Birgin2005}. Common penalties 
include $P_2$~\cite{KortBertsekas1976}, $P_3$~\cite{Nakayama1975} and PHR~\cite{Hestenes1969,Powell1969}. 
Following 
prior literature on ALM in the context of CNNs~\cite{rony2021augmented,liu2023cals}, we resort to the PHR function as the penalty, which is defined as:
\begin{equation}
\label{eq:phr}
    \mathrm{PHR}(z,\lambda,\rho) =
    \begin{cases}
        \lambda z + \frac12 \rho z^2 \quad &\text{if} \quad \lambda + \rho z \geq 0;\\
        -\frac{\lambda^2}{2\rho} \quad &\text{otherwise}.
    \end{cases}
\end{equation}
Our whole approach is detailed in Alg.~\ref{algo:ours} (\Cref{sec:algo}).

\begin{table*}
\centering
\scriptsize
\begin{tabular}{clcccccccccccc}
\toprule
\multirow{2}{*}{Score} & \multirow{2}{*}{Method} & \multicolumn{3}{c}{MNIST-LT} & \multicolumn{3}{c}{CIFAR10-LT} & \multicolumn{3}{c}{CIFAR100-LT} & \multicolumn{3}{c}{ImageNet-LT} \\
\cmidrule(lr){3-5} \cmidrule(lr){6-8} \cmidrule(lr){9-11} \cmidrule(lr){12-14} 
 &  & S & C & CG & S & C & CG & S & C & CG & S & C & CG \\
\midrule
\multirow[l]{8}{*}{THR} & CE & 4.32 & 0.90 & 6.93 & 6.65 & 0.90 & 6.65 & 8.30 & 0.90 & 6.90 & 9.25 & 0.90 & 5.32 \\
 & FL~\cite{lin2018focalloss} & 4.90 & 0.90 & 5.95 & 6.30 & 0.90 & 6.30 & 9.23 & 0.90 & 6.50 & 8.85 & 0.90 & 5.70 \\
 & ConfTr~\cite{stutz2022conftr}\textcolor{gray}{$_\text{ ICLR'22}$} & 4.35 & 0.90 & 4.92 & 5.05 & 0.90 & 6.29 & 7.84 & 0.90 & 6.31 & 6.60 & 0.90 & 5.64 \\
 & CUT~\cite{einbinder2022cut}\textcolor{gray}{$_\text{ NeurIPS'22}$} & 3.71 & 0.90 & 4.90 & 4.33 & 0.90 & 6.17 & 4.84 & 0.90 & 5.07 & \color{blue} \bfseries 5.38 & 0.90 & 5.42 \\
 & InfoCTr Fano~\cite{correia2025infocp}\textcolor{gray}{$_\text{ NeurIPS'24}$} & \color{blue} \bfseries 2.88 & 0.90 & 4.47 & 4.51 & 0.90 & 5.69 & 4.78 & 0.90 & 4.96 & 7.14 & 0.90 & 5.08 \\
 & DPSM~\cite{shi2025direct}\textcolor{gray}{$_\text{ ICML'25}$} & 3.06 & 0.90 & \color{blue} \bfseries 4.32 & \color{blue} \bfseries 3.65 & 0.90 & \color{blue} \bfseries 5.06 & \color{blue} \bfseries 4.21 & 0.90 & 4.91 & 5.78 & 0.90 & 4.71 \\
 & \cellcolor{Prune!20}\ours + HR (Ours) & \cellcolor{Prune!20}3.00 & \cellcolor{Prune!20}0.90 & \cellcolor{Prune!20}4.42 & \cellcolor{Prune!20}3.76 & \cellcolor{Prune!20}0.90 & \cellcolor{Prune!20}5.14 & \cellcolor{Prune!20}5.38 & \cellcolor{Prune!20}0.90 & \color{blue} \bfseries \cellcolor{Prune!20}4.58 & \cellcolor{Prune!20}5.72 & \cellcolor{Prune!20}0.90 & \color{blue} \bfseries \cellcolor{Prune!20}4.67 \\
 & \cellcolor{Prune!20}\ours + ALM (Ours) & \color{red} \bfseries \cellcolor{Prune!20}2.62 & \cellcolor{Prune!20}0.90 & \color{red} \bfseries \cellcolor{Prune!20}4.22 & \color{red} \bfseries \cellcolor{Prune!20}3.29 & \cellcolor{Prune!20}0.90 & \color{red} \bfseries \cellcolor{Prune!20}4.42 & \color{red} \bfseries \cellcolor{Prune!20}3.03 & \cellcolor{Prune!20}0.90 & \color{red} \bfseries \cellcolor{Prune!20}4.53 & \color{red} \bfseries \cellcolor{Prune!20}4.99 & \cellcolor{Prune!20}0.90 & \color{red} \bfseries \cellcolor{Prune!20}4.40 \\
\midrule
\multirow[l]{8}{*}{APS} & CE & 5.25 & 0.90 & 5.18 & 6.80 & 0.90 & 5.80 & 9.20 & 0.90 & 6.80 & 9.40 & 0.90 & 5.15 \\
 & FL~\cite{lin2018focalloss} & 5.48 & 0.90 & 4.72 & 6.45 & 0.90 & 5.45 & 9.85 & 0.90 & 6.07 & 9.54 & 0.90 & 5.64 \\
 & ConfTr~\cite{stutz2022conftr}\textcolor{gray}{$_\text{ ICLR'22}$} & 5.15 & 0.90 & 4.80 & 5.19 & 0.90 & 4.67 & 8.17 & 0.90 & 5.31 & 7.99 & 0.90 & 5.01 \\
 & CUT~\cite{einbinder2022cut}\textcolor{gray}{$_\text{ NeurIPS'22}$} & 5.03 & 0.90 & 4.26 & 3.94 & 0.90 & 4.56 & 6.97 & 0.90 & 4.55 & \color{blue} \bfseries 6.08 & 0.90 & 4.74 \\
 & InfoConfTr Fano~\cite{correia2025infocp}\textcolor{gray}{$_\text{ NeurIPS'24}$} & \color{blue} \bfseries 3.69 & 0.90 & \color{red} \bfseries 3.78 & 5.13 & 0.90 & \color{red} \bfseries 3.52 & 7.18 & 0.90 & 5.65 & 7.27 & 0.90 & \color{blue} \bfseries 4.18 \\
 & DPSM~\cite{shi2025direct}\textcolor{gray}{$_\text{ ICML'25}$} & 3.83 & 0.90 & 4.33 & \color{blue} \bfseries 3.55 & 0.90 & \color{blue} \bfseries 4.28 & 6.60 & 0.90 & 4.76 & 7.02 & 0.90 & 4.51 \\
 & \cellcolor{Prune!20}\ours + HR (Ours) & \cellcolor{Prune!20}4.02 & \cellcolor{Prune!20}0.90 & \cellcolor{Prune!20}4.34 & \cellcolor{Prune!20}3.62 & \cellcolor{Prune!20}0.90 & \cellcolor{Prune!20}4.29 & \color{blue} \bfseries \cellcolor{Prune!20}5.65 & \cellcolor{Prune!20}0.90 & \color{blue} \bfseries \cellcolor{Prune!20}4.12 & \cellcolor{Prune!20}6.97 & \cellcolor{Prune!20}0.90 & \cellcolor{Prune!20}4.62 \\
 & \cellcolor{Prune!20}\ours + ALM (Ours) & \color{red} \bfseries \cellcolor{Prune!20}3.53 & \cellcolor{Prune!20}0.90 & \color{blue} \bfseries \cellcolor{Prune!20}4.12 & \color{red} \bfseries \cellcolor{Prune!20}2.99 & \cellcolor{Prune!20}0.90 & \cellcolor{Prune!20}4.52 & \color{red} \bfseries \cellcolor{Prune!20}4.49 & \cellcolor{Prune!20}0.90 & \color{red} \bfseries \cellcolor{Prune!20}4.10 & \color{red} \bfseries \cellcolor{Prune!20}5.89 & \cellcolor{Prune!20}0.90 & \color{red} \bfseries \cellcolor{Prune!20}3.90 \\
\bottomrule
\end{tabular}

\caption{\textbf{Main results across long-tailed image classification datasets.} Prediction set size (S), empirical coverage (C) and coverage gap (CG) values
with $\alpha=0.1$ and an imbalance factor of $\gamma=0.1$ for all datasets except ImageNet-LT~\cite{liu2019imagenetlt}. The best model 
yields the smallest set size and coverage gap, while maintaining coverage close to the desired level. Best result in {\textbf{\color{red}Red}}, whereas {\textbf{\color{blue}Blue}} indicates the second best result. See~\Cref{sec:Additional Experimental Results}
for results with RAPS~\cite{angelopoulos2022raps}.}
\label{tab:results_gamma0.1_main}
\vspace{-3mm}
\end{table*}

\paragraph{Differentiability.} The original objective is non-smooth as it involves indicator functions to compute the size of the prediction set and requires computing the empirical batch-level quantile in~\eqref{eq:set-size-definition}. To make it differentiable, we will need to smooth the objective, as in~\cite{stutz2022conftr}.
In particular, we use the sigmoid function to smooth the indicator $\mathds{1}\left[a\leq b\right]$, defined as: $\widehat{\mathds{1}}\left[a\leq b\right] = 1 / (1 + \exp(-(b - a) / T))$ where $T > 0$ is a tunable hyper-parameter to control its smoothness.
Moreover, we compute the conformal threshold using a differentiable quantile operator, derived using smooth sorting techniques~\cite{grover2019neuralsort, petersen2022diffsort}. Smooth approximations relax the calibration step in Split CP, resulting in the loss of the marginal coverage guarantee.
However, as the temperature hyper-parameter $T \to 0$, the smooth formulation converges to the original non-smooth computations, and the corresponding coverage guarantee.
This is sufficient because the smoothing operations are replaced {\em only} during training to enable gradient-based optimization. At test time, we use the original non-smooth implementation, and the formal coverage guarantee follows directly~\cite{angelopoulos2022raps, romano2020aps, sadinle2018thr, angelopoulos2022intro}.

\section{Experiments and Results}

\subsection{Experimental Setup}

\paragraph{Datasets.} \ours is empirically validated in several computer vision classification tasks, including {MNIST} ~\cite{lecun1998mnist}, {CIFAR-10} and {CIFAR-100}~\cite{krizhevsky2009cifar}, {ImageNet}~\cite{russakovsky2015imagenet}, as well as {their long-tailed counterparts}~\cite{cao2019learning}. Furthermore, for experiments under imbalance distributions, we follow~\cite{cao2019learning} to collect the long-tailed versions of MNIST, CIFAR10 and CIFAR100 with different degrees of data imbalance ratio, $\gamma$, 
as well as {ImageNet-LT}~\cite{liu2019imagenetlt}. 
To show the general applicability of our approach, we further evaluate it on the {20 Newsgroups} dataset ~\cite{twenty_newsgroups_113}, a popular Natural Language Processing benchmark for text classification.

\paragraph{Conformity Scoring Functions.}
We consider three popular non-conformity scores: Least Ambiguous Classifier (THR)~\cite{sadinle2018thr}, Adaptive Prediction Sets (APS)~\cite{romano2020aps}, and Regularized Adaptive Prediction Sets (RAPS)~\cite{angelopoulos2022raps} (results for RAPS in~\Cref{sec:Additional Experimental Results}). 
Furthermore, unless otherwise stated, 
the mis-coverage level $\alpha=0.1$. 

\paragraph{Baselines.} We focus on learning-based methods:
\begin{enumerate*}[label=(\roman*)]
    \item CE, where we train the model with the standard cross-entropy loss;
    \item Focal loss (FL)~\cite{lin2018focalloss}, which mitigates the overconfident predictions 
    by down-weighting easy samples, \ie~with high probability;
    \item ConfTr~\cite{stutz2022conftr} minimizing the average differentiable prediction set size;
    \item CUT~\cite{einbinder2022cut} aiming at decreasing overconfidence by minimizing the gap between the uniform distribution and CDF of non-conformity scores;
    \item InfoCTr~\cite{correia2025infocp}, proposing three different upper bounds to the conditional entropy (Fano, Model-Based (MB) {Fano, and Data Processing Inequality (DPI)}) to achieve smaller average set size\footnote{We report results for the Fano bound and provide results for the MB Fano and DPI bounds in~\Cref{sec:Additional Experimental Results}.}; and
    \item DPSM~\cite{shi2025direct}, 
    which learns the quantile of the non-conformity scores instead of using a stochastic approximation.
\end{enumerate*}

\paragraph{Implementation Details.} We set the {smoothing factor} in FL to 3, and target size $\eta=1$, as in~\cite{stutz2022conftr}, for a fair comparison. 
For ALM hyper-parameters, we initialize $\vlamb^{(0)}=10^{-6}$, $\vrho^{(0)}=1$, $\beta=1.2$, and the penalty parameter $\vrho$ is updated every 10 epochs, as in~\cite{liu2023cals}.
In addition to using ALM to optimize for the multipliers (\ours + ALM), we consider an alternative approach, \ie, scaling them based on the value of the $k$-th penalty value $P_k^{(j)}$ for a multiplier $\lambda_k$ at iteration $j$: 
\begin{equation} \label{eq:heuristic}
\lambda_k^{(j+1)} = \begin{cases}
    \mu \cdot \lambda_k^{(j)} \quad & \quad \text{if } P_k^{(j+1)} > \tau P_k^{(j)},\\
    \lambda_k^{(j)} / \mu \quad & \quad \text{if } P_k^{(j)} > \tau P_k^{(j+1)},\\
    \lambda_k^{(j)} \quad & \quad \text{otherwise}. \nonumber\\
\end{cases}
\end{equation}
where $\mu, \tau>1$ are both fixed to $1.1$. We denote this heuristic rule as \ours + HR.

\begin{table*}
\centering
\scriptsize
\begin{tabular}{clccccccccccccccc}
\toprule
\multirow{2}{*}{Score} & \multirow{2}{*}{Method} & \multicolumn{3}{c}{MNIST} & \multicolumn{3}{c}{CIFAR10} & \multicolumn{3}{c}{CIFAR100} & \multicolumn{3}{c}{ImageNet} & \multicolumn{3}{c}{20 Newsgroups}\\
\cmidrule(lr){3-5} \cmidrule(lr){6-8} \cmidrule(lr){9-11} \cmidrule(lr){12-14} \cmidrule(lr){15-17} 
 &  & S & C & CG & S & C & CG & S & C & CG & S & C & CG & S & C & CG \\
\midrule
\multirow[l]{8}{*}{THR} & CE & 2.96 & 0.90 & 4.45 & 3.31 & 0.90 & 5.67 & 3.41 & 0.90 & 5.56 & 5.87 & 0.90 & 6.87 & 5.27 & 0.90 & 7.75 \\
 & FL~\cite{lin2018focalloss} & 3.89 & 0.90 & 3.33 & 3.15 & 0.90 & 6.78 & 4.05 & 0.90 & 4.98 & 5.75 & 0.90 & \color{blue} \bfseries 6.23 & 4.91 & 0.90 & 7.35 \\
 & ConfTr~\cite{stutz2022conftr}\textcolor{gray}{$_\text{ ICLR'22}$} & 3.45 & 0.90 & 4.99 & 2.89 & 0.90 & 5.08 & 4.76 & 0.90 & 5.65 & 5.45 & 0.90 & 6.41 & 4.34 & 0.90 & 7.27 \\
 & CUT~\cite{einbinder2022cut}\textcolor{gray}{$_\text{ NeurIPS'22}$} & 4.15 & 0.90 & 4.00 & 2.65 & 0.90 & 4.98 & 3.00 & 0.90 & 4.29 & 5.41 & 0.90 & 6.53 & 4.29 & 0.90 & 6.52 \\
 & InfoCTr Fano~\cite{correia2025infocp}\textcolor{gray}{$_\text{ NeurIPS'24}$} & 2.93 & 0.90 & 3.32 & \color{red} \bfseries 2.31 & 0.90 & \color{blue} \bfseries 4.37 & 4.12 & 0.90 & 4.24 & 5.01 & 0.90 & 6.50 & 4.36 & 0.90 & 6.19 \\
 & DPSM~\cite{shi2025direct}\textcolor{gray}{$_\text{ ICML'25}$} & 3.50 & 0.90 & \color{red} \bfseries 3.14 & 2.69 & 0.90 & 5.06 & 2.81 & 0.90 & \color{red} \bfseries 3.69 & \color{blue} \bfseries 5.00 & 0.90 & 6.36 & \color{blue} \bfseries 3.61 & 0.90 & 5.92 \\
 & \cellcolor{Prune!20}\ours + HR (Ours) & \color{blue} \bfseries \cellcolor{Prune!20}2.72 & \cellcolor{Prune!20}0.90 & \cellcolor{Prune!20}3.50 & \cellcolor{Prune!20}2.51 & \cellcolor{Prune!20}0.90 & \cellcolor{Prune!20}4.37 & \color{blue} \bfseries \cellcolor{Prune!20}2.69 & \cellcolor{Prune!20}0.90 & \cellcolor{Prune!20}4.34 & \cellcolor{Prune!20}5.86 & \cellcolor{Prune!20}0.90 & \cellcolor{Prune!20}6.74 & \cellcolor{Prune!20}3.72 & \cellcolor{Prune!20}0.90 & \color{blue} \bfseries \cellcolor{Prune!20}5.87 \\
 & \cellcolor{Prune!20}\ours + ALM (Ours) & \color{red} \bfseries \cellcolor{Prune!20}2.14 & \cellcolor{Prune!20}0.90 & \color{blue} \bfseries \cellcolor{Prune!20}3.21 & \color{blue} \bfseries \cellcolor{Prune!20}2.45 & \cellcolor{Prune!20}0.90 & \color{red} \bfseries \cellcolor{Prune!20}4.12 & \color{red} \bfseries \cellcolor{Prune!20}2.56 & \cellcolor{Prune!20}0.90 & \color{blue} \bfseries \cellcolor{Prune!20}4.21 & \color{red} \bfseries \cellcolor{Prune!20}4.99 & \cellcolor{Prune!20}0.90 & \color{red} \bfseries \cellcolor{Prune!20}6.20 & \color{red} \bfseries \cellcolor{Prune!20}3.59 & \cellcolor{Prune!20}0.90 & \color{red} \bfseries \cellcolor{Prune!20}5.74 \\
\midrule
\multirow[l]{8}{*}{APS} & CE & 3.46 & 0.90 & 3.86 & 3.41 & 0.90 & 3.56 & 5.50 & 0.90 & 4.09 & 7.21 & 0.90 & 6.54 & 5.35 & 0.90 & 6.90 \\
 & FL~\cite{lin2018focalloss} & 3.00 & 0.90 & 3.51 & 3.74 & 0.90 & \color{red} \bfseries 3.05 & 5.49 & 0.90 & 4.62 & 5.52 & 0.90 & \color{blue} \bfseries 6.15 & 5.09 & 0.90 & 6.55 \\
 & ConfTr~\cite{stutz2022conftr}\textcolor{gray}{$_\text{ ICLR'22}$} & 2.95 & 0.90 & 3.32 & 3.45 & 0.90 & 3.59 & \color{blue} \bfseries 4.57 & 0.90 & 3.82 & 5.95 & 0.90 & 6.42 & \color{blue} \bfseries 4.39 & 0.90 & 5.94 \\
 & CUT~\cite{einbinder2022cut}\textcolor{gray}{$_\text{ NeurIPS'22}$} & 2.71 & 0.90 & 3.78 & \color{red} \bfseries 2.76 & 0.90 & 3.47 & 4.87 & 0.90 & 3.80 & 5.89 & 0.90 & 7.70 & 4.54 & 0.90 & 6.12 \\
 & InfoConfTr Fano~\cite{correia2025infocp}\textcolor{gray}{$_\text{ NeurIPS'24}$} & 2.43 & 0.90 & \color{red} \bfseries 3.09 & 3.08 & 0.90 & 3.86 & 4.95 & 0.90 & 3.84 & 5.70 & 0.90 & 6.65 & 4.56 & 0.90 & 5.69 \\
 & DPSM~\cite{shi2025direct}\textcolor{gray}{$_\text{ ICML'25}$} & 2.81 & 0.90 & 3.20 & \color{blue} \bfseries 2.84 & 0.90 & \color{blue} \bfseries 3.42 & 4.77 & 0.90 & 3.44 & \color{red} \bfseries 4.56 & 0.90 & 6.24 & 4.74 & 0.90 & 5.40 \\
 & \cellcolor{Prune!20}\ours + HR (Ours) & \color{red} \bfseries \cellcolor{Prune!20}2.28 & \cellcolor{Prune!20}0.90 & \cellcolor{Prune!20}3.35 & \cellcolor{Prune!20}3.09 & \cellcolor{Prune!20}0.90 & \cellcolor{Prune!20}3.87 & \cellcolor{Prune!20}4.76 & \cellcolor{Prune!20}0.90 & \color{blue} \bfseries \cellcolor{Prune!20}3.40 & \cellcolor{Prune!20}5.09 & \cellcolor{Prune!20}0.90 & \cellcolor{Prune!20}6.95 & \cellcolor{Prune!20}4.87 & \cellcolor{Prune!20}0.90 & \color{blue} \bfseries \cellcolor{Prune!20}5.36 \\
 & \cellcolor{Prune!20}\ours + ALM (Ours) & \color{blue} \bfseries \cellcolor{Prune!20}2.32 & \cellcolor{Prune!20}0.90 & \color{blue} \bfseries \cellcolor{Prune!20}3.15 & \cellcolor{Prune!20}2.97 & \cellcolor{Prune!20}0.90 & \cellcolor{Prune!20}3.56 & \color{red} \bfseries \cellcolor{Prune!20}3.99 & \cellcolor{Prune!20}0.90 & \color{red} \bfseries \cellcolor{Prune!20}3.22 & \color{blue} \bfseries \cellcolor{Prune!20}5.01 & \cellcolor{Prune!20}0.90 & \color{red} \bfseries \cellcolor{Prune!20}5.80 & \color{red} \bfseries \cellcolor{Prune!20}4.00 & \cellcolor{Prune!20}0.90 & \color{red} \bfseries \cellcolor{Prune!20}4.67 \\
\bottomrule
\end{tabular}
\caption{\textbf{Main results across image and text classification datasets.} Prediction set size (S), empirical coverage (C) and coverage gap (CG) values
with $\alpha=0.1$. The best model 
yields the smallest set size and coverage gap, while maintaining coverage close to the desired level. Best result in {\textbf{\color{red}Red}}, whereas {\textbf{\color{blue}Blue}} indicates the second best result. See~\Cref{sec:Additional Experimental Results}
for results with RAPS~\cite{angelopoulos2022raps}.}
\label{tab:results_gamma0.1}
\vspace{-3mm}
\end{table*}

\paragraph{Evaluation Metrics.}
We report average set size (S), marginal coverage (C), and deviation of class-conditional coverage from the target coverage $1 -\alpha$ (CG)~\cite{ding2023clustercp}.

Additional details related to implementation, datasets, baselines, and evaluation metrics are provided in~\Cref{sec:Additional Experimental Setup Details}.

\subsection{Main Results}

\paragraph{\ours shows robustness to long-tailed distributions.}
We begin our empirical evaluation with experiments on long-tailed distributions, which serves as a natural and rigorous testbed for evaluating the advantages of learning per-class penalties. In such regimes, the data distribution is heavily imbalanced, with a small number of head classes containing abundant samples and a large number of tail classes with limited observations. To this end, we train on a long-tailed version of each dataset with an imbalance factor $\gamma = 0.1$, while keeping the calibration and test sets unchanged. This setup preserves exchangeability between the calibration and test samples, ensuring that the theoretical coverage guarantees of conformal prediction remain valid. \Cref{tab:results_gamma0.1_main} reports average set size and the coverage gap of 
THR and APS, across four widely used classification benchmarks and compared to baselines and related works. Existing conformal training methods fail to consistently improve efficiency on 
long-tail distributions, and even FL, designed for imbalance, produces large prediction sets. In contrast, \ours reduces average set size by learning class-conditional penalties from coverage gaps, improving efficiency across both THR and APS. In particular, \ours yields the best \textit{efficiency} (\ie, set size) across all datasets, while exhibiting a strong \textit{adaptivity}, as it achieves the lowest coverage gap in all but in case (MNIST-LT/APS), where it ranks second. Interestingly, the heuristic rule HR in~\eqref{eq:heuristic} remains competitive, often ranking second-best, further underscoring \ours’s effectiveness under the challenging class imbalance scenario.

\paragraph{From long-tailed to balanced distributions.} While long-tailed datasets highlight the challenge of class imbalance, it is equally important to assess performance on standard benchmarks with more balanced label distributions. This evaluation allows us to test whether \ours adapts beyond imbalance-specific scenarios and remains effective when classes are uniformly represented. By comparing results on widely used balanced datasets (Table \ref{tab:results_gamma0.1}), we demonstrate the versatility of \ours in shaping prediction sets across diverse regimes. \ours + ALM often ranks first and second across all five datasets and non-conformity scores. These gains are particularly important in datasets with large number of categories (\eg, CIFAR100 or ImageNet), or higher semantic complexity (\eg, 20 Newsgroups, where text classification involves high‑dimensional sparse features, semantic ambiguity, and nuanced language). Indeed, if we consider only these three datasets, the proposed approach presents a clear superiority over existing methods.

\paragraph{Varying imbalance factor $\gamma$.} 
To further investigate the robustness of the proposed class-specific penalty formulation, we conduct a controlled study across different values of the imbalance parameter $\gamma \in \{0.5, 0.8, 1.0\}$, which modulates the long-tailedness of the data distribution. Specifically, $\gamma$ determines the relative frequency decay between head and tail classes, with lower values corresponding to more severe imbalance. This analysis enables us to systematically assess how the proposed per-class penalty weights $\{\lambda_k\}_{k=1}^{|\mathcal{Y}|}$ adapt to varying degrees of distributional skew. As shown in~\Cref{tab:ablation_gamma_cifar100lt_main}, \ours demonstrates superior adaptivity and robustness across varying imbalance regimes. All methods struggle with inflated prediction sets, yet our approach maintains more compact sets with lower coverage gaps compared to other baselines. Notably, as in the case of extreme imbalance (\ie,~\Cref{tab:results_gamma0.1_main}), FL exhibits poor adaptation with substantially larger sets and higher coverage disparities across all $\gamma$ values. Even as the distribution becomes more balanced ($\gamma\rightarrow1.0$), \ours consistently learns to optimally shape class-conditional set sizes compared to other existing conformal training methods, yielding superior results.

\begin{table}[h!]
\centering
\resizebox{\linewidth}{!}{
\begin{tabular}{clcccccccccccc}
\toprule
& \multirow{2}{*}{Method} & \multicolumn{3}{c}{$\gamma=0.5$} & \multicolumn{3}{c}{$\gamma=0.8$} & \multicolumn{3}{c}{$\gamma=1.0$} \\
\cmidrule(lr){3-5} \cmidrule(lr){6-8} \cmidrule(lr){9-11} \cmidrule(lr){12-14}
 &  & S & C & CG & S & C & CG & S & C & CG \\
\midrule
\multirow[l]{8}{*}{\rotatebox[origin=c]{90}{THR}} & CE & 7.54 & 0.90 & 6.80 & 4.25 & 0.90 & 6.02 & 3.41 & 0.90 & 5.56 \\
 & FL~\cite{lin2018focalloss} & 6.86 & 0.90 & 5.98 & 4.46 & 0.90 & 5.57 & 4.05 & 0.90 & 4.98 \\
 & ConfTr~\cite{stutz2022conftr} & 6.43 & 0.90 & 6.01 & 4.94 & 0.90 & 5.69 & 4.76 & 0.90 & 5.65 \\
 & CUT~\cite{einbinder2022cut}& 4.02 & 0.90 & 4.89 & 3.80 & 0.90 & 4.62 & 3.00 & 0.90 & 4.29 \\
 & InfoCTr Fano~\cite{correia2025infocp}& 4.56 & 0.90 & 4.90 & 4.22 & 0.90 & \color{blue} \bfseries 4.30 & 4.12 & 0.90 & 4.24 \\
 & DPSM~\cite{shi2025direct}& \color{blue} \bfseries 3.72 & 0.90 & 4.77 & \color{blue} \bfseries 3.15 & 0.90 & 4.35 & 2.81 & 0.90 & \color{red} \bfseries 3.69 \\
 & \cellcolor{Prune!20}\ours + HR (Ours) & \cellcolor{Prune!20}4.30 & \cellcolor{Prune!20}0.90 & \color{blue} \bfseries \cellcolor{Prune!20}4.49 & \cellcolor{Prune!20}3.61 & \cellcolor{Prune!20}0.90 & \cellcolor{Prune!20}4.40 & \color{blue} \bfseries \cellcolor{Prune!20}2.69 & \cellcolor{Prune!20}0.90 & \cellcolor{Prune!20}4.34 \\
 & \cellcolor{Prune!20}\ours + ALM (Ours) & \color{red} \bfseries \cellcolor{Prune!20}2.88 & \cellcolor{Prune!20}0.90 & \color{red} \bfseries \cellcolor{Prune!20}4.37 & \color{red} \bfseries \cellcolor{Prune!20}2.73 & \cellcolor{Prune!20}0.90 & \color{red} \bfseries \cellcolor{Prune!20}4.29 & \color{red} \bfseries \cellcolor{Prune!20}2.56 & \cellcolor{Prune!20}0.90 & \color{blue} \bfseries \cellcolor{Prune!20}4.21 \\
\bottomrule
\end{tabular}
}
\caption{\textbf{Ablation study of imbalance factor $\gamma$ on CIFAR100-LT.} Performance comparison across different levels of class imbalance severity (lower $\gamma$ values indicate more severe imbalance). See~\Cref{sec:Additional Experimental Results} for results with APS~\cite{romano2020aps} and RAPS~\cite{angelopoulos2022raps}.} 
\label{tab:ablation_gamma_cifar100lt_main}
\end{table}

\begin{figure}[t]
    \centering
    \includegraphics[width=1\linewidth]{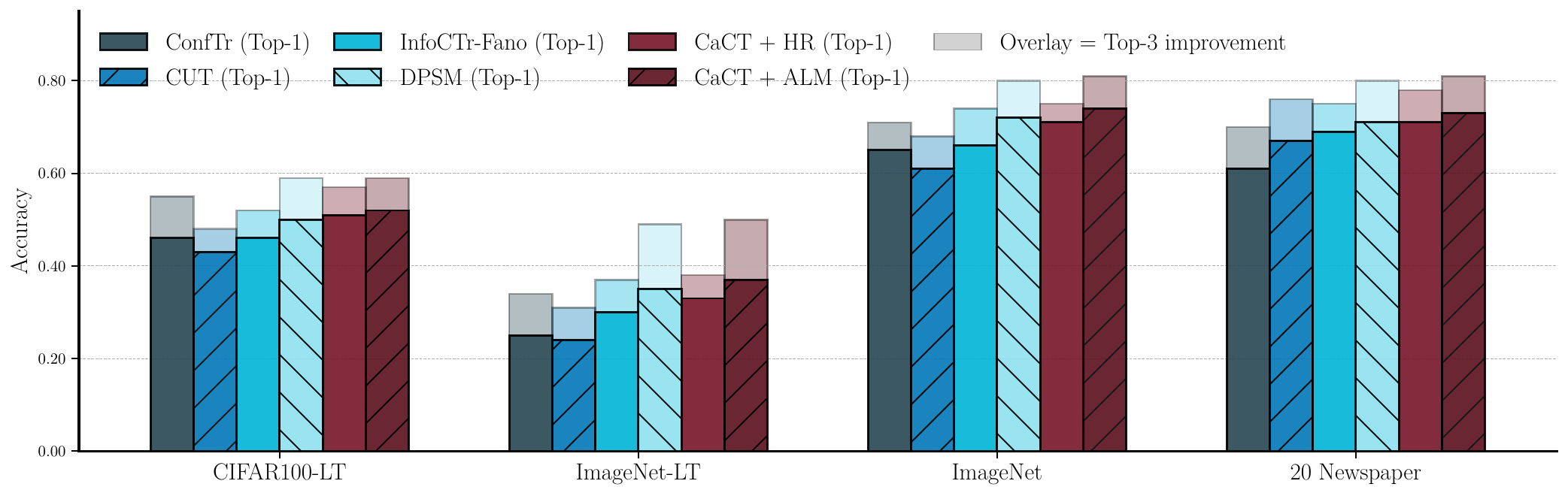}
    \caption{\textbf{Discriminative results of conformal training methods.} Top-$k$ ($k={1,3}$) results across several challenging datasets.}
    \label{fig:topk_main}
     \vspace{-3mm}
\end{figure}

\paragraph{Discriminative performance of \ours (\cref{fig:topk_main}).}
While conformal methods are primarily evaluated on coverage and set size, it is also important to assess their discriminative power, \ie, how often the true label appears among the top-ranked predictions. Across all datasets, \ours consistently ranks among the top performers in Top-1 accuracy, with \ours + ALM often outperforming prior methods, \eg, ConfTr, CUT, and InfoCTr Fano. The gains are particularly pronounced on ImageNet and 20 Newsgroups, where semantic complexity and class granularity pose greater challenges.

\paragraph{Varying mis-coverage level $\alpha$ at test-time.} In conformal training, we need to fix a target mis-coverage level to compute batch-level quantiles. In all of our experiments, we set $\alpha=0.01$. We now investigate whether the performance of the resulting models deteriorates at different coverage rates, potentially ``overfitting'' to the value of $\alpha$ used for training. In~\Cref{tab:ablation_alpha_cifar100lt}, we show how the average set size and coverage gap vary for different mis-coverage levels $\alpha$ at test time for models trained via conformal training using $\alpha=0.01$ and models trained with CE and FL, which are agnostic to the coverage rate. In all cases, \ours produces a better average set size and a better coverage gap for THR, APS, and RAPS, showing that \ours learns to optimally control the prediction set size even at varying mis-coverage levels.

\begin{table}[h!]
\centering
\resizebox{\linewidth}{!}{
\begin{tabular}{clcccccccccccc}
\toprule
& \multirow{2}{*}{Method} & \multicolumn{3}{c}{$\alpha=0.01$} & \multicolumn{3}{c}{$\alpha=0.05$} & \multicolumn{3}{c}{$\alpha=0.1$} \\
\cmidrule(lr){3-5} \cmidrule(lr){6-8} \cmidrule(lr){9-11} \cmidrule(lr){12-14}
 &  & S & C & CG & S & C & CG & S & C & CG \\
\midrule
\multirow[l]{8}{*}{\rotatebox[origin=c]{90}{THR}} & CE & 16.93 & 0.99 & 4.66 & 12.54 & 0.95 & 5.38 & 8.30 & 0.90 & 6.90 \\
 & FL~\cite{lin2018focalloss} & 18.71 & 0.99 & \color{blue} \bfseries 3.59 & 13.22 & 0.95 & 4.54 & 9.23 & 0.90 & 6.50 \\
 & ConfTr~\cite{stutz2022conftr}\textcolor{gray}{$_\text{ ICLR'22}$} & 20.34 & 0.99 & 4.06 & 11.78 & 0.95 & \color{blue} \bfseries 4.09 & 7.84 & 0.90 & 6.31 \\
 & CUT~\cite{einbinder2022cut}\textcolor{gray}{$_\text{ NeurIPS'22}$} & 14.26 & 0.99 & 3.67 & 9.01 & 0.95 & 4.12 & 4.84 & 0.90 & 5.07 \\
 & InfoCTr Fano~\cite{correia2025infocp}\textcolor{gray}{$_\text{ NeurIPS'24}$} & \color{blue} \bfseries 13.19 & 0.99 & 4.44 & 9.55 & 0.95 & 4.83 & 4.78 & 0.90 & 4.96 \\
 & DPSM~\cite{shi2025direct}\textcolor{gray}{$_\text{ ICML'25}$} & 14.87 & 0.99 & 4.78 & 8.58 & 0.95 & 4.84 & \color{blue} \bfseries 4.21 & 0.90 & 4.91 \\
 & \cellcolor{Prune!20}\ours + HR (Ours) & \cellcolor{Prune!20}14.69 & \cellcolor{Prune!20}0.99 & \cellcolor{Prune!20}4.32 & \color{blue} \bfseries \cellcolor{Prune!20}8.54 & \cellcolor{Prune!20}0.95 & \cellcolor{Prune!20}4.57 & \cellcolor{Prune!20}5.38 & \cellcolor{Prune!20}0.90 & \color{blue} \bfseries \cellcolor{Prune!20}4.58 \\
 & \cellcolor{Prune!20}\ours + ALM (Ours) & \color{red} \bfseries \cellcolor{Prune!20}12.08 & \cellcolor{Prune!20}0.99 & \color{red} \bfseries \cellcolor{Prune!20}3.40 & \color{red} \bfseries \cellcolor{Prune!20}7.21 & \cellcolor{Prune!20}0.95 & \color{red} \bfseries \cellcolor{Prune!20}4.07 & \color{red} \bfseries \cellcolor{Prune!20}3.03 & \cellcolor{Prune!20}0.90 & \color{red} \bfseries \cellcolor{Prune!20}4.53 \\
\bottomrule
\end{tabular}
}
\caption{\textbf{Ablation study of test-time mis-coverage level $\alpha$ on CIFAR100-LT}. Lower is better. {\textbf{\color{red}Red}}: best result; {\textbf{\color{blue}Blue}}: second best. See~\Cref{sec:Additional Experimental Results} for results with APS~\cite{romano2020aps} and RAPS~\cite{angelopoulos2022raps}.}
\label{tab:ablation_alpha_cifar100lt}
\end{table}

\begin{figure}
    \centering
\includegraphics[width=1\linewidth]{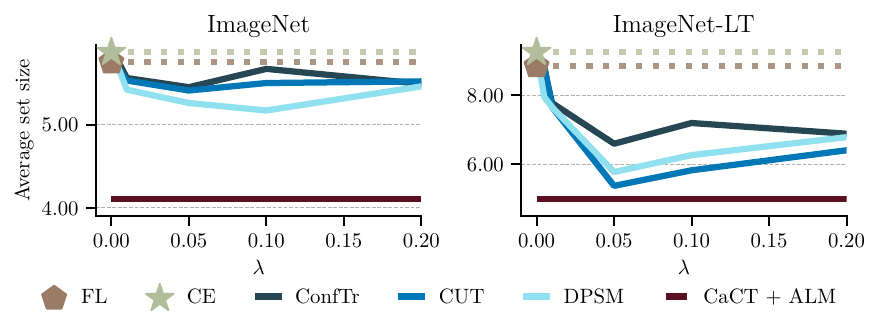}
    \caption{\textbf{Sensitivity to the balancing term $\lambda$.}
    Selecting an appropriate $\lambda$ for existing methods, \eg, ConfTr~\cite{stutz2022conftr} and CUT~\cite{einbinder2022cut} is challenging, especially on complex datasets like ImageNet-LT.}
    \label{fig:lambda_main}
    \vspace{-4mm}
\end{figure}

\paragraph{\ours is agnostic to the choice of the penalty weight.}
As shown in~\Cref{fig:class-conditional-size-coverage}, when a single global penalty weight is applied across all classes, as in ConfTr, CUT, and InfoCTr, the resulting conformal sets tend to be biased toward overrepresented classes. This bias leads to suboptimal coverage and large prediction sets for underrepresented categories. Introducing class-dependent penalty weights allows the model to adjust the regularization strength in a data-adaptive manner, thereby achieving more balanced coverage and more compact prediction sets across the distribution. Furthermore, using the ALM to adaptively learn these penalty weights, as shown in~\Cref{fig:lambda_main}, enables \ours to produce smaller and more efficient prediction sets compared to other conformal training methods with a tuned scalar weight.

\begin{figure}[h!]
    \centering
    \includegraphics[width=\columnwidth]{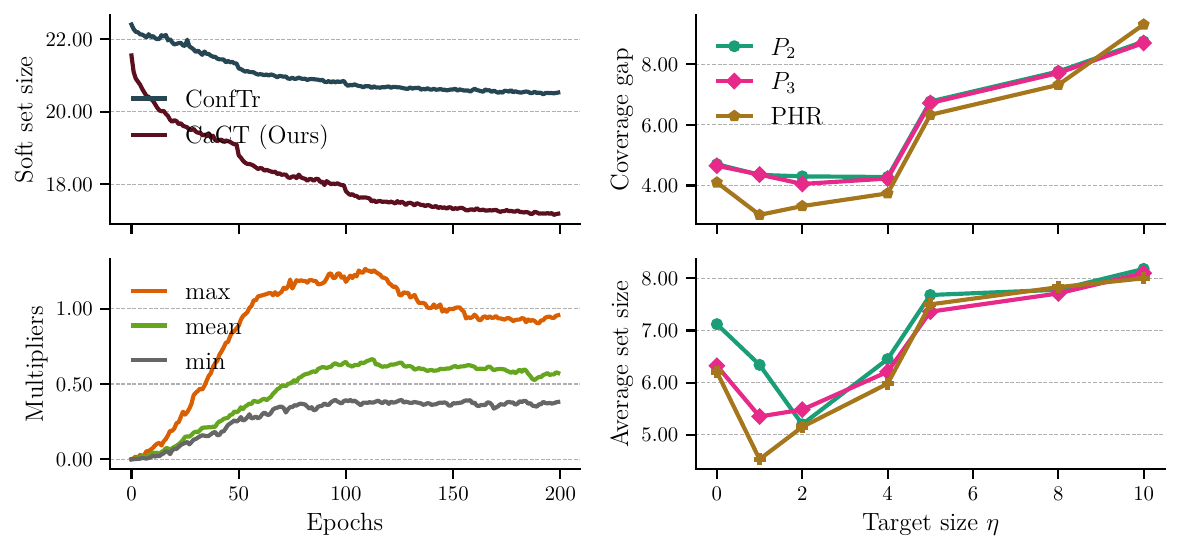}
    \caption{\textbf{Left: Evolution of average soft set size} (approximated using Sigmoid function) of \ours and ConfTr (\textit{top}), and values of multipliers $\vlamb$ for \ours after each training epoch (\textit{bottom}). \textbf{Right: Effect of penalty functions and target size}: Coverage gap and average set size on test set are shown across three popular penalty functions and different target size values. {Ablation study on CIFAR100-LT} using ResNet and THR score.}
    \label{fig:ablation_alm}
    \vspace{-3mm}
\end{figure}

\paragraph{Convergence Analysis.}
In Fig.~\ref{fig:ablation_alm}, we ablate the role of the target size $\eta$ and evaluate the impact of three commonly used penalty functions, $P_2$, $P_3$, and $\mathrm{PHR}$, for ALM 
(see App.~\ref{sec:penalties_axioms} for details). 
\textit{Left figures} illustrate the evolution of the Lagrange multipliers $\vlamb$ and average set size over training epochs, where we observe two distinct training regimes. Initially, the multipliers are large, enforcing strong constraints and leading to larger prediction sets. As training progresses, both the multipliers and the average set size decrease, indicating that the model has learned to satisfy the constraints and thus requires less penalization. \textit{Right figure} compares performance across the three penalty functions with different values of target set size $\eta$. We find that $\mathrm{PHR}$ 
yields the best results, and setting $\eta = 1.0$ (as in \cite{stutz2022conftr}) achieves the best trade-off between prediction set size and coverage gap.

\section{Conclusion}
In this paper, we introduced \ours, a novel class adaptive conformal training approach. 
We reformulate the conformal training problem using a modified Augmented Lagrangian Multiplier algorithm, enabling adaptive learning of class-specific constraints. Extensive experiments demonstrate that \ours can effectively shape prediction sets class-conditionally, reducing average set size for 
underrepresented classes and improving the efficiency of existing conformal prediction methods, such as THR~\cite{sadinle2018thr} and APS~\cite{romano2020aps}, and outperforming prior conformal training methods like ConfTr~\cite{stutz2022conftr}, or DPSM~\cite{shi2025direct},  
even under severe class imbalance.


{\small\bibliographystyle{ieeenat_fullname}\bibliography{main}}









\appendix
\maketitlesupplementary



\section{Additional Background Details}
\label{sec:Additional Background Details}

\subsection{Non-conformity scoring functions}
Non-conformity scores are measures of the predictive model's uncertainty. They quantify the level of agreement between the input and its corresponding output, with higher scores indicating greater uncertainty. These scores are a fundamental component of conformal prediction, and many advancements in the field stem from the development of new scoring functions. In the following, we present three widely used scoring functions in the literature.

\paragraph{Least Ambiguous Classifier.}
The least ambiguous classifier (THR)~\cite{sadinle2018thr} is the simplest and most commonly used non-conformity function. It yields the smallest possible prediction sets by only considering the confidence assigned to the true label. The score is defined as
\begin{equation}
    S^\mathrm{THR}_\theta(X, Y) = 1 - \pi_{\theta}(X)_Y,
\end{equation}
where $\pi_{\theta}(x)_y$ is the predicted probability for the true label $y$.
Despite its efficiency, it tends to produce empty prediction sets when the model is highly uncertain, and undercover hard subgroups and overcover easy ones~\cite{angelopoulos2022intro}.

\paragraph{Adaptive Prediction Sets.}
To address this issue, adaptive prediction sets (APS)~\cite{romano2020aps} were proposed. APS aims to handle ambiguity by considering the entire distribution over classes rather than focusing solely on the true label. The intuition behind APS is to build prediction sets by sequentially including labels in descending order of their predictive probabilities until the true label is covered. This approach ensures that the prediction set adapts to the model's confidence level and avoids having empty sets. If the model assigns high probability to the true label, the resulting set is small; otherwise, it grows accordingly. Formally, the APS score is defined as:
\begin{multline}
    S^\mathrm{APS}_\theta(X, Y) =
    \sum_{i=1}^{r_\theta(X, Y)-1} \pi_{\theta}(X)_i\\
    + U \cdot \pi_{\theta}(X)_{r_\theta(X, Y)},
\end{multline}
where $r_\theta$ gives the rank of label $y$ among the predictive probabilities $\pi_{\theta}(x)$ sorted in descending order, and $U\in[0, 1]$ is a uniform random variable added to achieve exact marginal coverage.

\paragraph{Regularized Adaptive Prediction Sets.}
Building on APS, the regularized adaptive prediction sets (RAPS)~\cite{angelopoulos2022raps} introduce a regularization term that discourages the inclusion of too many low-confidence labels.
This function encourages smaller prediction sets by penalizing cases where the true label is not among the top-$k$ candidates.
The RAPS score is given by:
\begin{multline}
    S^\mathrm{RAPS}_\theta(X, Y) = 
    S^\mathrm{APS}_\theta(X, Y) \\
    + \lambda_\mathrm{reg} \cdot \max \left( 0, r_\theta(X, Y) - k \right),
\end{multline}
where $\lambda_\mathrm{reg}$ is a regularization coefficient that controls the strength of the penalty, and $k$ is the desired target rank.

\subsection{Conformal Training Objectives}
\label{sec:Objective function for conformal training methods}

ConfTr~\cite{stutz2022conftr} estimates a soft measure of the average prediction set size, defined as:
\begin{equation}
    \widehat{\mathcal{L}}_{s, \mathrm{ConfTr}} (\theta) =
    \frac{1}{n} \sum_{i=1}^n \sum_{y\in\mathcal{Y}} \widehat{\mathds{1}} \left[ S_\theta(X_i,y) \leq \widehat{q}_\theta \right]
\end{equation}
CUT~\cite{einbinder2022cut} measures the maximum deviation from uniformity of the non-conformity scores, defined as,
\begin{equation}
    \widehat{\mathcal{L}}_{s, \mathrm{CUT}} (\theta) = \sup_{w \in [0, 1]} \left| \widehat{F}_S(w) - w \right|,
\end{equation}
where $\widehat{F}_S(w)$ is the empirical cumulative distribution function (CDF) of $S_i=S_\theta(X_i,Y_i)$ for $i \in [n]$. CUT could also be written with respect to the quantiles as follows,
\begin{equation}
    \sup_{\alpha \in [0, 1]} \left| (1-\alpha) - \widehat{q}_\theta \right|
\end{equation}
InfoCTr~\cite{correia2025infocp} proposes three different ways to upper bound the conditional entropy
$H[Y | X]$, which can serve as better alternatives to push the model closer to the true distribution and achieve a smaller average set size. \citet{correia2025infocp} propose:
\begin{enumerate*}[label=(\roman*)]
    \item a bound derived from the data processing inequality (DPI bound), and
    \item two bounds based on a variation of Fano's inequality~\cite{fano1961transmission}: a model-agnostic version (simple Fano bound) and a version informed by the predictive model itself (model-based Fano bound).
\end{enumerate*}

ConfTr, CUT, and InfoCTr use a stochastic approximation to estimate quantiles. DPSM~\cite{shi2025direct} proposes a bi-level optimization approach for conformal training, where the quantile value is learned by minimizing the pinball loss on non-conformity scores. This serves as a constraint to achieve accurate quantiles while simultaneously optimizing for efficiency, similar to the objective used in ConfTr.

\subsection{Conditional Conformal Prediction}
\label{sec:Conditional Conformal Prediction}
Standard Conformal Prediction (CP), \ie, Split CP, provides marginal coverage guarantees, meaning that the prediction set contains the true label with probability at least $1-\alpha$ on average over the data distribution.
However, this guarantee may fail to protect minority groups and rare but important classes.

\paragraph{Label-Conditional CP}
guarantees that each ground-truth class $y$ has at least $1-\alpha$ probability of being included in the prediction set~\cite{vovk2012conditional, vovk2003mcp}.
It computes a separate conformal threshold for each class in the calibration set, and the prediction set is defined as
\begin{equation}
    \mathcal{C}_\theta(X) = \left\{ y \in \mathcal{Y} \mid S_\theta(X,y) \leq \widehat{q}_\theta^y \right\},
\end{equation}
where the threshold $\widehat{q}_\theta^y$ for label $y$ is given by
\begin{equation}
    \widehat{q}_\theta^y = \mathcal{Q}\left( \{S_\theta(X_i,Y_i)\}_{i\in\mathcal{I}_y}; \frac{\lceil (n_y + 1)(1 - \alpha) \rceil}{n_y} \right),
\end{equation}
and $n_y$ is the number of calibration samples belonging to class $y$, with $\mathcal{I}_y$ denoting their indices.
If a class contains too few samples to compute a threshold (\ie, 
$n_y < (1/\alpha) - 1$, we set $\widehat{q}_\theta^y=\infty$, which would include the entire label space $\mathcal{Y}$ in the prediction set.
LabelCP may still suffer from uninformative and overly conservative prediction sets (\ie, large sets) for classes with limited calibration samples~\cite{ding2023clustercp}.

\paragraph{Cluster-Conditional CP} outperforms Split CP and LabelCP, in terms of class-conditional coverage, when there is limited access to calibration data per-class~\cite{ding2023clustercp}.
ClusterCP groups the classes into $M$ clusters based on their non-conformity scores and then computes a shared conformal threshold for each cluster:
\begin{equation}
    \mathcal{C}_\theta(X) = \left\{ y \in \mathcal{Y} \mid S_\theta(X,y) \leq \widehat{q}_\theta^{h(y)} \right\},
\end{equation}
where $h : \mathcal{Y} \to \{1, \dots, M\} \cup \{\mathrm{null}\}$ is a clustering function (\eg, K-Means).
The threshold for each cluster is given by
\begin{align}
    &\widehat{q}_\theta^k = \mathcal{Q}\left( \{S_\theta(X_i,Y_i)\}_{i\in\mathcal{I}_k}; \frac{\lceil (n_k + 1)(1 - \alpha) \rceil}{n_k} \right),\\
    &\widehat{q}_\theta^\mathrm{null} = \mathcal{Q}\left( \{S_\theta(X_i,Y_i)\}_{i\in\mathcal{I}}; \frac{\lceil (n + 1)(1 - \alpha) \rceil}{n} \right),    
\end{align}
where $\mathcal{I}_k = \{i \mid h(Y_i)=k\}$ and $n_k=|\mathcal{I}_k|$ denote the indices and the number of calibration samples in cluster $k$, respectively.
If a class has too few samples to be assigned to any cluster, it is placed in the $\mathrm{null}$ group, for which the conformal threshold is computed using all calibration samples.

In addition to the experiments we have carried out in the main manuscript, we assess the performance of conformal training objectives with LabelCP and ClusterCP in~\Cref{tab:labelcp_gamma0.1,tab:clustercp_gamma0.1}.

\section{Penalty Functions for ALM}
\label{sec:penalties_axioms}

We outline here the requirements for a penalty function $P$ in the Augmented Lagrangian Method (ALM), detailed in~\Cref{ssec:our_method}.

A function $P : \real \times \real_{++} \times \real_{++}  \to  \real $ is considered a valid a Penalty-Lagrangian function if it satisfies the following properties:
\begin{enumerate*}[label=(\roman*)]
    \item \textit{Non-negativity:} $P(z, \lambda, \rho) \geq 0$ for all $z \in \mathbb{R}, \lambda, \rho \in \mathbb{R}_{++}$, and
    \item \textit{Regularity:} $P$ is continuous in all arguments, and its partial derivative with respect to $z$, denoted $P'(z, \lambda, \rho) \equiv \partial P(z, \lambda, \rho) / {\partial z}$, exists and is continuous over the same domain.
\end{enumerate*}
In addition, a penalty function $P$ should satisfy the following four axioms~\cite{Birgin2005}:
\begin{enumerate}[label={\bf Axiom \arabic*.}]
    \item $P'(z, \lambda, \rho) \geq 0 \quad \forall\, z\in\real, \lambda, \rho \in \real_{++}^2$.
    \item $P'(0, \lambda, \rho) = \lambda  \quad \forall\, \lambda, \rho \in \real_{++}^2$.
    \item If, for all $j\in\N, \; \lambda^{(j)} \in [\lambda_\text{min},\lambda_\text{max}]$, where $0 < \lambda_\text{min} \leq \lambda_\text{max} < \infty$, then: $\lim\limits_{j \to \infty}\rho^{(j)}=\infty$ and $\lim\limits_{j \to \infty}z^{(j)}>0$ imply that $\lim\limits_{j \to \infty}P'(z^{(j)}, \lambda^{(j)}, \rho^{(j)})=\infty$.
    \item If, for all $j\in\N, \; \lambda^{(j)} \in [\lambda_\text{min},\lambda_\text{max}]$, where $0 < \lambda_\text{min} \leq \lambda_\text{max} < \infty$, then: $\lim\limits_{j \to \infty}\rho^{(j)}=\infty$ and $\lim\limits_{j \to \infty}z^{(j)}<0$ imply that $\lim\limits_{j \to \infty}P'(z^{(j)}, \lambda^{(j)}, \rho^{(j)})=0$.
\end{enumerate}

The first two axioms guarantee that the derivative of the Penalty-Lagrangian function $P$ \wrt $z$ is positive and equals to $\lambda$ when $z=0$. The last two axioms guarantee that the derivative tends to infinity when the constraint is not satisfied, and zero otherwise. 
There exist in the literature many penalty functions. in this paper we use the, in addition to, PHR function and empirically compare it with two other penalty functions: $P_2$ and $P_3$, given by
\begin{align}
    \mathrm{P_2}(z, \lambda, \rho) &= 
    \begin{cases}
        \lambda z + \lambda \rho z^2 + \frac{1}{6} \rho^2 z^3 & \text{if } z \geq 0, \\
        \frac{\lambda z}{1 - \rho z} & \text{otherwise},
    \end{cases} \\
    \mathrm{P_3}(z, \lambda, \rho) &= 
    \begin{cases}
        \lambda z + \lambda \rho z^2 & \text{if } z \geq 0, \\
        \frac{\lambda z}{1 - \rho z} & \text{otherwise}.
    \end{cases}
\end{align}

\begin{figure}[!htbp]
    \centering
    \includegraphics[width=\linewidth]{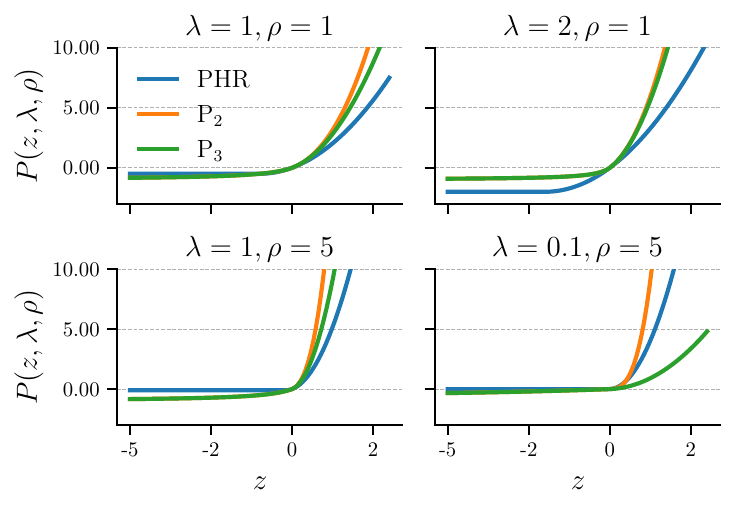}
    \caption{\textbf{Examples of penalty functions, $P_2$, $P_3$, and $\mathrm{PHR}$ for different values of $\lambda$ and $\rho$}. The validity of such functions is discusses in~\Cref{sec:penalties_axioms}.}
    \label{fig:illustr-penalty-functs}
\end{figure}

\section{\ours Algorithm}
\label{sec:algo}
Our approach is detailed in~\Cref{algo:ours}, which describes the proposed class adaptive training procedure to efficiently shape the class-conditional set size distribution across categories.
\misc{\textit{Model Optimization:}} In the \textit{inner} step, the model parameters $\theta$ are updated via stochastic gradient descent. For each mini-batch drawn from the training set $\mathcal{D}_\mathrm{train}$, the data is randomly partitioned into a calibration subset $\mathcal{B}_\mathrm{cal}$ and a prediction subset $\mathcal{B}_\mathrm{pred}$. The main learning objective, which incorporates both the conformal objective and penalty terms, is evaluated on these subsets. Then, the gradient of this loss with respect to $\theta$ is used to perform a parameter update. Note that this step corresponds to solving the primal problem under the current penalty configuration.
\misc{\textit{Multiplier and Penalty Updates:}} Then, during the \textit{outer} iteration, the algorithm adjusts the penalty multipliers $\boldsymbol{\lambda}$ and penalty parameters $\boldsymbol{\rho}$ for each class $k \in \mathcal{Y}$. This reflects the dual ascent mechanism commonly used in ALM frameworks.

\begin{algorithm}[h]
    \caption{Class Adaptive Conformal Training with ALM}
    \label{algo:ours}
    \small
    \begin{algorithmic}[1]
        \Require
        Training set $\mathcal{D}_\mathrm{train}$,
        validation set $\mathcal{D}_\mathrm{val}$,
        learning rate $\nu >0$,
        mis-coverage level $\alpha \in [0, 1]$,
        training epochs $T$,
        target size $\eta \geq 0$,
        penalty parameter increase rate $\beta > 1$,
        penalty function $P$,
        initial multiplier $\vlamb^{(0)}$,
        initial penalty parameter $\vrho^{(0)}$
        \For{$j=1, \dots, T$}
            \markcomment{1}{$\triangleright$ Inner Step \misc{(Model optimization)}}
            \For{each mini-batch $\mathcal{B}$ from $\mathcal{D}_\mathrm{train}$}
                \State Randomly split $\mathcal{B}$ into $\mathcal{B}_\mathrm{cal}$ and $\mathcal{B}_\mathrm{pred}$
                \State Compute $\widehat{q}_\theta$ as the $(1-\alpha)$-quantile of non-conformity scores on $\mathcal{B}_\mathrm{cal}$
                \State $\widehat{\nabla}_{\theta}^{(j)} \gets \nabla_\theta \widehat{\mathcal{L}}_\mathrm{\ours}(\theta^{(j)}, \vlamb^{(j)}, \vrho^{(j)})$ on $\mathcal{B}_\mathrm{pred}$
                \State $\theta^{(j+1)} \gets \theta^{(j)} + \nu \cdot \widehat{\nabla}_{\theta}^{(j)}$
            \EndFor

            \markcomment{1}{$\triangleright$ Outer Step \misc{(Multiplier and Penalty Updates)}}
            \For{$k = 1, \dots, |\mathcal{Y}|$}
                \State $\widehat{d}_k \gets \frac{1}{n_k}\sum_{i\in\mathcal{I}_k} |\mathcal{C}_{\theta}(X_i)|$ \footnotemark
                \State $\lambda_{k}^{(j+1)} \gets P^\prime \left( \frac{\widehat{d}_k}{\eta} - 1, \lambda_{k}^{(j)}, \rho_{k}^{(j)} \right)$
                \If{the $k$-th constraint does not improve~(\cref{eq:rho_update_rule})}
                    \State{$\rho_{k}^{(j+1)} \gets \beta \cdot \rho_{k}^{(j+1)}$}
                \EndIf
            \EndFor
        \EndFor
        \Ensure Model parameters $\theta^{(T)}$
    \end{algorithmic}
\end{algorithm}
\footnotetext{$\mathcal{I}_k=\{i \in \mathcal{D}_\mathrm{val} \mid Y_i=y\}$ is the set of indices of validation samples that belong to class $k \in \mathcal{Y}$, and $n_k=|\mathcal{I}_k|$.}

\section{Additional Experimental Setup Details}
\label{sec:Additional Experimental Setup Details}

\paragraph{Dataset and Split.}
All datasets come with pre-defined training/test splits. We also experiment with long-tailed datasets: manually resampled long-tailed versions of MNIST, CIFAR10 and CIFAR100 with an exponential decay of $\gamma$ (\ie, $\gamma=\min_i n_i / \max_i n_i$, where $n_i$ is the number of samples in each class), an imbalance factor following~\cite{cao2019learning}, and ImageNet-LT~\cite{liu2019imagenetlt} which is truncated from ImageNet~\cite{russakovsky2015imagenet} by sampling a subset so that the labels of the training set follow a long-tailed distribution. In our experiments, we further split the test set 50:50 into a validation subset and a final test set $\mathcal{D}_\mathrm{test}$. The validation subset is then further divided 20:80 into a validation set $\mathcal{D}_\mathrm{val}$ (used for ALM optimization) and a calibration set $\mathcal{D}_\mathrm{cal}$. This final partition ensures that $\mathcal{D}_\mathrm{cal}$ remains exchangeable with the test samples; reusing the same data for both ALM updates and calibration would violate exchangeability. This mimics the proceedure followed in other conformal training approaches, such as \cite{stutz2022conftr}.

\paragraph{Models and Training.}
We use a linear model on MNIST, ResNet-34/50~\cite{he2015resnet} on CIFAR10 and CIFAR100, ResNet-50 for ImageNet, and a pre-trained BERT~\cite{devlin2019bert} model with linear layer on top of it for 20 Newsgroups.
For CIFAR datasets, since the default ResNet implementation produces a $1 \times 1$ feature map before global pooling, we modify the first convolution to use a kernel size of 3, stride 1, and padding 1.
We train using stochastic gradient descent (SGD) with momentum set to 0.9 and Nesterov gradients. The learning rate followed a step-wise schedule, where the initial learning rate was multiplied by 0.1 at $2/5$, $3/5$, and $4/5$ of the total number of training epochs.
all models are trained for 200 epochs, with the exception of MNIST and 20 Newsgroups dataset which were trained only for 50 epochs.
The learning rate and batch size are optimized alongside the hyper-parameters using grid search, see next paragraph.
We only use data augmentations on the CIFAR and ImageNet datasets, and following~\cite{correia2025infocp},
we only apply random flipping and cropping.
Note that the same thing applies to the corresponding long-tailed version of each dataset.

\paragraph{Numerical Stability.}
To mitigate numerical instabilities in ALM, often encountered with non-linear penalties, in practice we normalize the constraints using the target size parameter $\eta > 0$. Specifically, the original constraint $\widehat{d}_k - \eta \leq 0$ is equivalently rewritten as ${\widehat{d}_k}/{\eta} - 1 \leq 0$. This normalization improves the numerical behavior of the ALM updates and ensures more stable convergence during training~\cite{liu2023cals}.

\paragraph{Choice of Non-Conformity Score During Training.}
We perform conformal training by using the log probabilities as model outputs rather than the THR or APS scores. This choice follows the findings of \citet{stutz2022conftr} and is consistent with the observations of \citet{correia2025infocp} and \citet{shi2025direct}. We found this approach to be efficient and to improve stability during training. The models generalize well across all non-conformity score functions evaluated at test time, which confirms the robustness of this design choice.

\paragraph{Fine-tuning CIFAR and ImageNet.}
Base models are first trained on the full training set using cross-entropy loss, and then fine-tuned using the different conformal training approaches: ConfTr~\cite{stutz2022conftr}, CUT~\cite{einbinder2022cut}, InfoCTr~\cite{correia2025infocp}, DPSM~\cite{shi2025direct}, and our approach \ours. During fine-tuning, the final (logit) layer is reinitialized and trained using the same data augmentation as in the base training phase.
For ImageNet, the base model is a pre-trained ResNet-50 available in Torchvision~\cite{torchvision2016}. 

\paragraph{Hyper-Parameters.}
The final hyper-parameters were obtained using grid search over the following hyper-parameters:
batch size in $\{100, 500, 1000\}$;
learning rate in $\{0.05, 0.01, 0.005\}$;
temperature parameter $T \in \{0.01, 0.1, 0.5, 1\}$;
regularization weight $\lambda \in \{0.001, 0.005, 0.01, 0.05, 0.1, 0.2\}$, used in ConfTr~\cite{stutz2022conftr}, CUT~\cite{einbinder2022cut}, InfoCTr~\cite{correia2025infocp}, and DPSM~\cite{shi2025direct} to control the trade-off between the classification loss and regularization term;
steepness, which controls the smoothness of the differentiable sorting algorithm~\cite{petersen2022diffsort} in $\{1, 10, 100\}$. Higher values make the output closer to standard (hard) sorting.
Following~\cite{stutz2022conftr}, hyper-parameter optimization is done using the training dataset. For each hyper-parameter configuration, we evaluate the average prediction set size across 10 random calibration/test splits, using THR at test time. We select the hyper-parameters that yield the smallest prediction set size.

\paragraph{Evaluation metrics.} We formally define the metrics introduced in the empirical validation. Using the test set $\mathcal{D}_\mathrm{test}$ across all datasets, we report:
\begin{itemize}
    \item The empirical coverage to measure the degree of satisfaction of the marginal coverage guarantees. It is defined as:
    \begin{equation}
        \mathrm{Cov} = \frac{1}{|\mathcal{D}_\mathrm{test}|} \sum_{i\in\mathcal{D}_\mathrm{test}} \mathds{1}\!\left[Y_i \in \mathcal{C}_\theta(X_i)\right].\nonumber
    \end{equation}
    \item The average set size, also known as inefficiency~\cite{stutz2022conftr}, is a widely employed metric for assessing the utility of CP methods for multi-class classification problems. An optimum conformal method in terms of efficiency should provide a lower set size. It is formally defined as:
    \begin{equation}
        \mathrm{Size} = \frac{1}{|\mathcal{D}_\mathrm{test}|} \sum_{i\in\mathcal{D}_\mathrm{test}} \left|\mathcal{C}_\theta(X_i)\right|.
    \end{equation}
    \item The deviation of class-conditional coverage from the desired coverage level $1-\alpha$, also known as coverage gap~\cite{ding2023clustercp}. It is defined as
    \begin{equation}
        \mathrm{CovGap} = 100 \times \frac{1}{|\mathcal{Y}|} \sum_{y\in\mathcal{Y}} | c_y - (1-\alpha) |,
    \end{equation}
    where $c_y = \frac{1}{n_y} \sum_{i\in\mathcal{I}_y} \mathds{1}\!\left[ Y_i \in \mathcal{C}_\theta(X_i) \right]$, and $n_y$ is the number of samples in the dataset for class $y$.
    \item Top-$k$ accuracy, to assess the discriminative performance of the underlying predictive model. For each input, the model produces class scores that determine a ranked list of labels. Top-$k$ accuracy measures the proportion of test examples for which the true label appears among the $k$ highest scoring predictions.
\end{itemize}

\section{Additional Experimental Results}
\label{sec:Additional Experimental Results}


\paragraph{Performance across the three versions of InfoCTr~\cite{correia2025infocp}.}
\citet{correia2025infocp} introduced three bounds, namely Fano, MB Fano and DPI, which we review in~\Cref{sec:Objective function for conformal training methods}. For clarity in our empirical study, we reported in the main paper only the results obtained with the Fano bound because it consistently delivered the strongest performance among the three. To support this choice, \Cref{tab:infoConf_LT,tab:infoConf_std} provide the complete comparison of all three bounds against \ours. The results show that the Fano bound achieves the strongest performance among the three bounds, which aligns with the findings of~\citet{correia2025infocp}.
This behavior is expected because the Fano bound provides a tighter lower bound on the set size.
In~\Cref{tab:infoConf_LT}, for the \textbf{long tailed setting}, where the number of training samples decays exponentially across classes, our method achieves the best results for every dataset. Furthermore, when using RAPS as non-conformity score, we also observe that the heuristic rule for optimizing the multipliers performs second best after the ALM version, which further supports the robustness of our approach.
In~\Cref{tab:infoConf_std}, which reports the results for the \textbf{balanced setting}, where all classes have the same number of training samples, \ours remains the strongest or second strongest method across all image classification benchmarks. This holds regardless of the choice of non conformity score for Split CP. Furthermore, \ours performs competitively on the text classification dataset, yielding the best results across the three non-conformity scores, highlighting the versatility of our method and showing that it is not limited to vision based tasks.

\begin{table*}[!htbp]
\centering
\scriptsize
\begin{tabular}{clcccccccccccc}
\toprule
\multirow{2}{*}{Score} & \multirow{2}{*}{Method} & \multicolumn{3}{c}{MNIST-LT} & \multicolumn{3}{c}{CIFAR10-LT} & \multicolumn{3}{c}{CIFAR100-LT} & \multicolumn{3}{c}{ImageNet-LT} \\
\cmidrule(lr){3-5} \cmidrule(lr){6-8} \cmidrule(lr){9-11} \cmidrule(lr){12-14}
 &  & S & C & CG & S & C & CG & S & C & CG & S & C & CG \\
\midrule
\multirow[l]{5}{*}{THR} & InfoCTr Fano~\cite{correia2025infocp}\textcolor{gray}{$_\text{ NeurIPS'24}$} & \color{blue} \bfseries 2.88 & 0.90 & 4.47 & 4.51 & 0.90 & 5.69 & \color{blue} \bfseries 4.78 & 0.90 & 4.96 & 7.14 & 0.90 & 5.08 \\
 & InfoCTr MB Fano~\cite{correia2025infocp}\textcolor{gray}{$_\text{ NeurIPS'24}$} & 3.05 & 0.90 & 5.28 & 4.93 & 0.90 & \color{blue} \bfseries 5.08 & 7.02 & 0.90 & 5.11 & 7.11 & 0.90 & \color{blue} \bfseries 4.66 \\
 & InfoCTr DPI~\cite{correia2025infocp}\textcolor{gray}{$_\text{ NeurIPS'24}$} & 2.95 & 0.90 & 4.45 & 5.38 & 0.90 & 5.72 & 7.32 & 0.90 & 6.05 & 6.68 & 0.90 & 4.99 \\
 & \cellcolor{Prune!20}\ours + HR (Ours) & \cellcolor{Prune!20}3.00 & \cellcolor{Prune!20}0.90 & \color{blue} \bfseries \cellcolor{Prune!20}4.42 & \color{blue} \bfseries \cellcolor{Prune!20}3.76 & \cellcolor{Prune!20}0.90 & \cellcolor{Prune!20}5.14 & \cellcolor{Prune!20}5.38 & \cellcolor{Prune!20}0.90 & \color{blue} \bfseries \cellcolor{Prune!20}4.58 & \color{blue} \bfseries \cellcolor{Prune!20}5.72 & \cellcolor{Prune!20}0.90 & \cellcolor{Prune!20}4.67 \\
 & \cellcolor{Prune!20}\ours + ALM (Ours) & \color{red} \bfseries \cellcolor{Prune!20}2.62 & \cellcolor{Prune!20}0.90 & \color{red} \bfseries \cellcolor{Prune!20}4.22 & \color{red} \bfseries \cellcolor{Prune!20}3.29 & \cellcolor{Prune!20}0.90 & \color{red} \bfseries \cellcolor{Prune!20}4.42 & \color{red} \bfseries \cellcolor{Prune!20}3.03 & \cellcolor{Prune!20}0.90 & \color{red} \bfseries \cellcolor{Prune!20}4.53 & \color{red} \bfseries \cellcolor{Prune!20}4.99 & \cellcolor{Prune!20}0.90 & \color{red} \bfseries \cellcolor{Prune!20}4.40 \\
\midrule
\multirow[l]{5}{*}{APS} & InfoCTr Fano~\cite{correia2025infocp}\textcolor{gray}{$_\text{ NeurIPS'24}$} & \color{blue} \bfseries 3.69 & 0.90 & \color{red} \bfseries 3.78 & 5.13 & 0.90 & \color{red} \bfseries 3.52 & 7.18 & 0.90 & 5.65 & 7.27 & 0.90 & \color{blue} \bfseries 4.18 \\
 & InfoCTr MB Fano~\cite{correia2025infocp}\textcolor{gray}{$_\text{ NeurIPS'24}$} & 4.80 & 0.90 & 4.65 & 3.77 & 0.90 & 4.91 & 7.36 & 0.90 & 4.46 & \color{blue} \bfseries 6.63 & 0.90 & 4.32 \\
 & InfoCTr DPI~\cite{correia2025infocp}\textcolor{gray}{$_\text{ NeurIPS'24}$} & 4.56 & 0.90 & \color{blue} \bfseries 4.06 & 5.30 & 0.90 & 4.48 & 7.24 & 0.90 & 4.97 & 8.10 & 0.90 & 4.61 \\
 & \cellcolor{Prune!20}\ours + HR (Ours) & \cellcolor{Prune!20}4.02 & \cellcolor{Prune!20}0.90 & \cellcolor{Prune!20}4.34 & \color{blue} \bfseries \cellcolor{Prune!20}3.62 & \cellcolor{Prune!20}0.90 & \color{blue} \bfseries \cellcolor{Prune!20}4.29 & \color{blue} \bfseries \cellcolor{Prune!20}5.65 & \cellcolor{Prune!20}0.90 & \color{blue} \bfseries \cellcolor{Prune!20}4.12 & \cellcolor{Prune!20}6.97 & \cellcolor{Prune!20}0.90 & \cellcolor{Prune!20}4.62 \\
 & \cellcolor{Prune!20}\ours + ALM (Ours) & \color{red} \bfseries \cellcolor{Prune!20}3.53 & \cellcolor{Prune!20}0.90 & \cellcolor{Prune!20}4.12 & \color{red} \bfseries \cellcolor{Prune!20}2.99 & \cellcolor{Prune!20}0.90 & \cellcolor{Prune!20}4.52 & \color{red} \bfseries \cellcolor{Prune!20}4.49 & \cellcolor{Prune!20}0.90 & \color{red} \bfseries \cellcolor{Prune!20}4.10 & \color{red} \bfseries \cellcolor{Prune!20}5.89 & \cellcolor{Prune!20}0.90 & \color{red} \bfseries \cellcolor{Prune!20}3.90 \\
\midrule
\multirow[l]{5}{*}{RAPS} & InfoCTr Fano~\cite{correia2025infocp}\textcolor{gray}{$_\text{ NeurIPS'24}$} & 4.21 & 0.90 & 4.29 & 6.68 & 0.90 & 4.89 & \color{blue} \bfseries 7.21 & 0.90 & \color{blue} \bfseries 4.29 & 7.39 & 0.90 & 5.04 \\
 & InfoCTr MB Fano~\cite{correia2025infocp}\textcolor{gray}{$_\text{ NeurIPS'24}$} & 3.39 & 0.90 & 4.48 & 5.60 & 0.90 & 5.34 & 9.09 & 0.90 & 4.99 & 6.78 & 0.90 & 5.07 \\
 & InfoCTr DPI~\cite{correia2025infocp}\textcolor{gray}{$_\text{ NeurIPS'24}$} & 4.86 & 0.90 & 4.77 & 6.25 & 0.90 & 5.01 & 8.84 & 0.90 & 4.83 & 5.92 & 0.90 & 4.98 \\
 & \cellcolor{Prune!20}\ours + HR (Ours) & \color{blue} \bfseries \cellcolor{Prune!20}3.34 & \cellcolor{Prune!20}0.90 & \color{blue} \bfseries \cellcolor{Prune!20}3.84 & \color{blue} \bfseries \cellcolor{Prune!20}4.05 & \cellcolor{Prune!20}0.90 & \color{blue} \bfseries \cellcolor{Prune!20}4.84 & \cellcolor{Prune!20}7.53 & \cellcolor{Prune!20}0.90 & \cellcolor{Prune!20}5.06 & \color{blue} \bfseries \cellcolor{Prune!20}5.65 & \cellcolor{Prune!20}0.90 & \color{blue} \bfseries \cellcolor{Prune!20}4.24 \\
 & \cellcolor{Prune!20}\ours + ALM (Ours) & \color{red} \bfseries \cellcolor{Prune!20}3.07 & \cellcolor{Prune!20}0.90 & \color{red} \bfseries \cellcolor{Prune!20}3.78 & \color{red} \bfseries \cellcolor{Prune!20}3.43 & \cellcolor{Prune!20}0.90 & \color{red} \bfseries \cellcolor{Prune!20}4.72 & \color{red} \bfseries \cellcolor{Prune!20}6.22 & \cellcolor{Prune!20}0.90 & \color{red} \bfseries \cellcolor{Prune!20}4.01 & \color{red} \bfseries \cellcolor{Prune!20}5.61 & \cellcolor{Prune!20}0.90 & \color{red} \bfseries \cellcolor{Prune!20}3.84 \\

\bottomrule
\end{tabular}
\caption{\textbf{Extension of~\Cref{tab:results_gamma0.1_main} -- Comparison between \ours and InfoCTr~\cite{correia2025infocp} bounds across long-tailed image classification datasets.} Prediction set size (S), empirical coverage (C) and coverage gap (CG) values
with $\alpha=0.1$ and an imbalance factor of $\gamma=0.1$ for all datasets except ImageNet-LT~\cite{liu2019imagenetlt}. The best model 
yields the smallest set size and coverage gap, while maintaining coverage close to the desired level. Best result in {\textbf{\color{red}Red}}, whereas {\textbf{\color{blue}Blue}} indicates the second best result.} 
\label{tab:infoConf_LT}
\end{table*}

\begin{table*}[!htbp]
\centering
\resizebox{\textwidth}{!}{
\begin{tabular}{clccccccccccccccc}
\toprule
\multirow{2}{*}{Score} & \multirow{2}{*}{Method} & \multicolumn{3}{c}{MNIST} & \multicolumn{3}{c}{CIFAR10} & \multicolumn{3}{c}{CIFAR100} & \multicolumn{3}{c}{ImageNet} & \multicolumn{3}{c}{20 Newsgroups} \\
\cmidrule(lr){3-5} \cmidrule(lr){6-8} \cmidrule(lr){9-11} \cmidrule(lr){12-14} \cmidrule(lr){15-17}
 &  & S & C & CG & S & C & CG & S & C & CG & S & C & CG & S & C & CG\\
\midrule
\multirow[l]{5}{*}{THR} & InfoCTr Fano~\cite{correia2025infocp}\textcolor{gray}{$_\text{ NeurIPS'24}$} & 2.93 & 0.90 & \color{blue} \bfseries 3.32 & \color{red} \bfseries 2.31 & 0.90 & \color{blue} \bfseries 4.37 & 4.12 & 0.90 & \color{blue} \bfseries 4.24 & \color{blue} \bfseries 5.01 & 0.90 & 6.50 & 4.36 & 0.90 & 6.19 \\
 & InfoCTr MB Fano~\cite{correia2025infocp}\textcolor{gray}{$_\text{ NeurIPS'24}$} & 3.00 & 0.90 & 3.40 & 2.98 & 0.90 & 5.39 & 3.49 & 0.90 & 4.56 & 5.49 & 0.90 & \color{blue} \bfseries 6.33 & 4.51 & 0.90 & 6.46 \\
 & InfoCTr DPI~\cite{correia2025infocp}\textcolor{gray}{$_\text{ NeurIPS'24}$} & 3.17 & 0.90 & 3.35 & 2.71 & 0.90 & 4.79 & 3.68 & 0.90 & 4.74 & 5.10 & 0.90 & 6.45 & 4.29 & 0.90 & 7.14 \\
 & \cellcolor{Prune!20}\ours + HR (Ours) & \color{blue} \bfseries \cellcolor{Prune!20}2.72 & \cellcolor{Prune!20}0.90 & \cellcolor{Prune!20}3.50 & \cellcolor{Prune!20}2.51 & \cellcolor{Prune!20}0.90 & \cellcolor{Prune!20}4.37 & \color{blue} \bfseries \cellcolor{Prune!20}2.69 & \cellcolor{Prune!20}0.90 & \cellcolor{Prune!20}4.34 & \cellcolor{Prune!20}5.86 & \cellcolor{Prune!20}0.90 & \cellcolor{Prune!20}6.74 & \color{blue} \bfseries \cellcolor{Prune!20}3.72 & \cellcolor{Prune!20}0.90 & \color{blue} \bfseries \cellcolor{Prune!20}5.87 \\
 & \cellcolor{Prune!20}\ours + ALM (Ours) & \color{red} \bfseries \cellcolor{Prune!20}2.14 & \cellcolor{Prune!20}0.90 & \color{red} \bfseries \cellcolor{Prune!20}3.21 & \color{blue} \bfseries \cellcolor{Prune!20}2.45 & \cellcolor{Prune!20}0.90 & \color{red} \bfseries \cellcolor{Prune!20}4.12 & \color{red} \bfseries \cellcolor{Prune!20}2.56 & \cellcolor{Prune!20}0.90 & \color{red} \bfseries \cellcolor{Prune!20}4.21 & \color{red} \bfseries \cellcolor{Prune!20}4.99 & \cellcolor{Prune!20}0.90 & \color{red} \bfseries \cellcolor{Prune!20}6.20 & \color{red} \bfseries \cellcolor{Prune!20}3.59 & \cellcolor{Prune!20}0.90 & \color{red} \bfseries \cellcolor{Prune!20}5.74 \\
\midrule
\multirow[l]{5}{*}{APS} & InfoCTr Fano~\cite{correia2025infocp}\textcolor{gray}{$_\text{ NeurIPS'24}$} & 2.43 & 0.90 & \color{red} \bfseries 3.09 & 3.08 & 0.90 & 3.86 & 4.95 & 0.90 & 3.84 & 5.70 & 0.90 & 6.65 & 4.56 & 0.90 & 5.69 \\
 & InfoCTr MB Fano~\cite{correia2025infocp}\textcolor{gray}{$_\text{ NeurIPS'24}$} & 2.56 & 0.90 & 3.29 & \color{blue} \bfseries 2.98 & 0.90 & \color{blue} \bfseries 3.75 & 4.93 & 0.90 & 3.72 & 5.29 & 0.90 & \color{blue} \bfseries 6.55 & \color{blue} \bfseries 4.44 & 0.90 & 5.50 \\
 & InfoCTr DPI~\cite{correia2025infocp}\textcolor{gray}{$_\text{ NeurIPS'24}$} & 2.62 & 0.90 & 3.16 & 3.19 & 0.90 & 3.84 & 5.01 & 0.90 & 3.83 & 5.18 & 0.90 & 6.60 & 4.68 & 0.90 & 6.75 \\
 & \cellcolor{Prune!20}\ours + HR (Ours) & \color{red} \bfseries \cellcolor{Prune!20}2.28 & \cellcolor{Prune!20}0.90 & \cellcolor{Prune!20}3.35 & \cellcolor{Prune!20}3.09 & \cellcolor{Prune!20}0.90 & \cellcolor{Prune!20}3.87 & \color{blue} \bfseries \cellcolor{Prune!20}4.76 & \cellcolor{Prune!20}0.90 & \color{blue} \bfseries \cellcolor{Prune!20}3.40 & \color{blue} \bfseries \cellcolor{Prune!20}5.09 & \cellcolor{Prune!20}0.90 & \cellcolor{Prune!20}6.95 & \cellcolor{Prune!20}4.87 & \cellcolor{Prune!20}0.90 & \color{blue} \bfseries \cellcolor{Prune!20}5.36 \\
 & \cellcolor{Prune!20}\ours + ALM (Ours) & \color{blue} \bfseries \cellcolor{Prune!20}2.32 & \cellcolor{Prune!20}0.90 & \color{blue} \bfseries \cellcolor{Prune!20}3.15 & \color{red} \bfseries \cellcolor{Prune!20}2.97 & \cellcolor{Prune!20}0.90 & \color{red} \bfseries \cellcolor{Prune!20}3.56 & \color{red} \bfseries \cellcolor{Prune!20}3.99 & \cellcolor{Prune!20}0.90 & \color{red} \bfseries \cellcolor{Prune!20}3.22 & \color{red} \bfseries \cellcolor{Prune!20}5.01 & \cellcolor{Prune!20}0.90 & \color{red} \bfseries \cellcolor{Prune!20}5.80 & \color{red} \bfseries \cellcolor{Prune!20}4.00 & \cellcolor{Prune!20}0.90 & \color{red} \bfseries \cellcolor{Prune!20}4.67 \\
\midrule
\multirow[l]{5}{*}{RAPS} & InfoCTr Fano~\cite{correia2025infocp}\textcolor{gray}{$_\text{ NeurIPS'24}$} & 3.13 & 0.90 & \color{red} \bfseries 3.08 & \color{blue} \bfseries 3.40 & 0.90 & 3.54 & \color{blue} \bfseries 3.45 & 0.90 & 3.41 & 5.78 & 0.90 & \color{blue} \bfseries 5.79 & 4.67 & 0.90 & \color{blue} \bfseries 4.68 \\
 & InfoCTr MB Fano~\cite{correia2025infocp}\textcolor{gray}{$_\text{ NeurIPS'24}$} & 3.19 & 0.90 & 3.26 & 3.51 & 0.90 & 3.46 & 4.76 & 0.90 & \color{blue} \bfseries 3.28 & 5.82 & 0.90 & 5.87 & 4.86 & 0.90 & 4.99 \\
 & InfoCTr DPI~\cite{correia2025infocp}\textcolor{gray}{$_\text{ NeurIPS'24}$} & \color{red} \bfseries 3.03 & 0.90 & 3.15 & 4.09 & 0.90 & \color{blue} \bfseries 3.44 & 4.85 & 0.90 & 3.31 & \color{blue} \bfseries 5.68 & 0.90 & 5.84 & 4.65 & 0.90 & 6.22 \\
 & \cellcolor{Prune!20}\ours + HR (Ours) & \cellcolor{Prune!20}3.75 & \cellcolor{Prune!20}0.90 & \cellcolor{Prune!20}3.31 & \cellcolor{Prune!20}4.19 & \cellcolor{Prune!20}0.90 & \cellcolor{Prune!20}3.63 & \cellcolor{Prune!20}4.19 & \cellcolor{Prune!20}0.90 & \cellcolor{Prune!20}3.78 & \cellcolor{Prune!20}5.83 & \cellcolor{Prune!20}0.90 & \cellcolor{Prune!20}6.86 & \color{blue} \bfseries \cellcolor{Prune!20}4.57 & \cellcolor{Prune!20}0.90 & \cellcolor{Prune!20}5.07 \\
 & \cellcolor{Prune!20}\ours + ALM (Ours) & \color{blue} \bfseries \cellcolor{Prune!20}3.09 & \cellcolor{Prune!20}0.90 & \color{blue} \bfseries \cellcolor{Prune!20}3.10 & \color{red} \bfseries \cellcolor{Prune!20}3.06 & \cellcolor{Prune!20}0.90 & \color{red} \bfseries \cellcolor{Prune!20}3.21 & \color{red} \bfseries \cellcolor{Prune!20}3.22 & \cellcolor{Prune!20}0.90 & \color{red} \bfseries \cellcolor{Prune!20}3.18 & \color{red} \bfseries \cellcolor{Prune!20}5.50 & \cellcolor{Prune!20}0.90 & \color{red} \bfseries \cellcolor{Prune!20}5.10 & \color{red} \bfseries \cellcolor{Prune!20}3.83 & \cellcolor{Prune!20}0.90 & \color{red} \bfseries \cellcolor{Prune!20}4.60 \\

\bottomrule
\end{tabular}
}
\caption{\textbf{Extension of~\Cref{tab:results_gamma0.1} -- Comparison between \ours and InfoCTr~\cite{correia2025infocp} bounds across image and text classification datasets.} Prediction set size (S), empirical coverage (C) and coverage gap (CG) values
with $\alpha=0.1$. The best model 
yields the smallest set size and coverage gap, while maintaining coverage close to the desired level. Best result in {\textbf{\color{red}Red}}, whereas {\textbf{\color{blue}Blue}} indicates the second best result.} 
\label{tab:infoConf_std}
\end{table*}

\paragraph{Results with RAPS~\cite{angelopoulos2022raps}.}
Due to length constraints, \Cref{tab:results_gamma0.1_main,tab:results_gamma0.1} in main paper present results only for THR and APS as non-conformity scores. To complement these results, \Cref{tab:results_gamma0.1_appendix_LT,tab:results_gamma0.1_appendix} report the performance of all methods on long-tailed and balanced datasets, respectively, when RAPS is used at testing. The results exhibit a performance trend similar to that observed for THR and APS. Across all conformal training objectives, \ours{} typically attains superior performance and shows greater robustness across datasets compared with alternative approaches.

\begin{table*}[!htbp]
\centering
\footnotesize
\begin{tabular}{clcccccccccccc}
\toprule
\multirow{2}{*}{Score} & \multirow{2}{*}{Method} & \multicolumn{3}{c}{MNIST-LT} & \multicolumn{3}{c}{CIFAR10-LT} & \multicolumn{3}{c}{CIFAR100-LT} & \multicolumn{3}{c}{ImageNet-LT} \\
\cmidrule(lr){3-5} \cmidrule(lr){6-8} \cmidrule(lr){9-11} \cmidrule(lr){12-14}
 &  & S & C & CG & S & C & CG & S & C & CG & S & C & CG \\
\midrule
\multirow[l]{8}{*}{RAPS} & CE & 5.21 & 0.90 & 4.04 & 6.87 & 0.90 & 6.12 & 9.15 & 0.90 & 6.12 & 9.56 & 0.90 & 5.23 \\
 & FL~\cite{lin2018focalloss} & 5.33 & 0.90 & 4.80 & 6.50 & 0.90 & 6.50 & 9.85 & 0.90 & 6.60 & 9.10 & 0.90 & 4.90 \\
 & ConfTr~\cite{stutz2022conftr}\textcolor{gray}{$_\text{ ICLR'22}$} & 4.69 & 0.90 & 4.27 & 5.67 & 0.90 & 5.87 & 9.23 & 0.90 & 5.16 & 8.26 & 0.90 & 4.85 \\
 & CUT~\cite{einbinder2022cut}\textcolor{gray}{$_\text{ NeurIPS'22}$} & 3.88 & 0.90 & \color{red} \bfseries 3.74 & 5.02 & 0.90 & 5.48 & 8.03 & 0.90 & 4.90 & \color{red} \bfseries 4.76 & 0.90 & 4.64 \\
 & InfoCTr Fano~\cite{correia2025infocp}\textcolor{gray}{$_\text{ NeurIPS'24}$} & 4.21 & 0.90 & 4.29 & 6.68 & 0.90 & 4.89 & 7.21 & 0.90 & \color{blue} \bfseries 4.29 & 7.39 & 0.90 & 5.04 \\
 & DPSM~\cite{shi2025direct}\textcolor{gray}{$_\text{ ICML'25}$} & 3.42 & 0.90 & \color{blue} \bfseries 3.77 & 4.21 & 0.90 & \color{red} \bfseries 4.71 & \color{blue} \bfseries 7.18 & 0.90 & 4.87 & 5.76 & 0.90 & 4.30 \\
 & \cellcolor{Prune!20}\ours + HR (Ours) & \color{blue} \bfseries \cellcolor{Prune!20}3.34 & \cellcolor{Prune!20}0.90 & \cellcolor{Prune!20}3.84 & \color{blue} \bfseries \cellcolor{Prune!20}4.05 & \cellcolor{Prune!20}0.90 & \cellcolor{Prune!20}4.84 & \cellcolor{Prune!20}7.53 & \cellcolor{Prune!20}0.90 & \cellcolor{Prune!20}5.06 & \cellcolor{Prune!20}5.65 & \cellcolor{Prune!20}0.90 & \color{blue} \bfseries \cellcolor{Prune!20}4.24 \\
 & \cellcolor{Prune!20}\ours + ALM (Ours) & \color{red} \bfseries \cellcolor{Prune!20}3.07 & \cellcolor{Prune!20}0.90 & \cellcolor{Prune!20}3.78 & \color{red} \bfseries \cellcolor{Prune!20}3.43 & \cellcolor{Prune!20}0.90 & \color{blue} \bfseries \cellcolor{Prune!20}4.72 & \color{red} \bfseries \cellcolor{Prune!20}6.22 & \cellcolor{Prune!20}0.90 & \color{red} \bfseries \cellcolor{Prune!20}4.01 & \color{blue} \bfseries \cellcolor{Prune!20}5.61 & \cellcolor{Prune!20}0.90 & \color{red} \bfseries \cellcolor{Prune!20}3.84 \\
\bottomrule
\end{tabular}
\caption{\textbf{Extension of~\Cref{tab:results_gamma0.1_main} -- Results across long-tailed image classification datasets (with RAPS \cite{angelopoulos2022raps} as non-conformity score).} Prediction set size (S), empirical coverage (C) and coverage gap (CG) values
with $\alpha=0.1$ and an imbalance factor of $\gamma=0.1$ for all datasets except ImageNet-LT~\cite{liu2019imagenetlt}. The best model 
yields the smallest set size and coverage gap, while maintaining coverage close to the desired level. Best result in {\textbf{\color{red}Red}}, whereas {\textbf{\color{blue}Blue}} indicates the second best result.} 
\label{tab:results_gamma0.1_appendix_LT}
\end{table*}

\begin{table*}[!htbp]
\centering
\footnotesize
\resizebox{\textwidth}{!}{
\begin{tabular}{clccccccccccccccc}
\toprule
\multirow{2}{*}{Score} & \multirow{2}{*}{Method} & \multicolumn{3}{c}{MNIST} & \multicolumn{3}{c}{CIFAR10} & \multicolumn{3}{c}{CIFAR100} & \multicolumn{3}{c}{ImageNet} & \multicolumn{3}{c}{20 Newsgroups} \\
\cmidrule(lr){3-5} \cmidrule(lr){6-8} \cmidrule(lr){9-11} \cmidrule(lr){12-14} \cmidrule(lr){15-17}
 &  & S & C & CG & S & C & CG & S & C & CG & S & C & CG & S & C & CG\\
\midrule
\multirow[l]{8}{*}{RAPS} & CE & 3.64 & 0.90 & 3.22 & 3.48 & 0.90 & 3.94 & 4.62 & 0.90 & 3.37 & 6.13 & 0.90 & 6.77 & 5.11 & 0.90 & 6.12 \\
 & FL~\cite{lin2018focalloss} & 3.39 & 0.90 & 3.18 & 3.98 & 0.90 & \color{blue} \bfseries 3.32 & 4.68 & 0.90 & 3.63 & 6.27 & 0.90 & 5.67 & 5.01 & 0.90 & 6.80 \\
 & ConfTr~\cite{stutz2022conftr}\textcolor{gray}{$_\text{ ICLR'22}$} & 3.26 & 0.90 & 3.28 & 3.18 & 0.90 & 3.49 & 4.30 & 0.90 & 3.83 & 6.35 & 0.90 & \color{blue} \bfseries 5.33 & 4.26 & 0.90 & 5.88 \\
 & CUT~\cite{einbinder2022cut}\textcolor{gray}{$_\text{ NeurIPS'22}$} & 3.63 & 0.90 & 3.84 & 3.12 & 0.90 & 3.92 & 4.55 & 0.90 & \color{red} \bfseries 3.16 & \color{red} \bfseries 5.47 & 0.90 & 5.84 & \color{blue} \bfseries 4.04 & 0.90 & 5.58 \\
 & InfoCTr Fano~\cite{correia2025infocp}\textcolor{gray}{$_\text{ NeurIPS'24}$} & \color{blue} \bfseries 3.13 & 0.90 & \color{red} \bfseries 3.08 & 3.40 & 0.90 & 3.54 & \color{blue} \bfseries 3.45 & 0.90 & 3.41 & 5.78 & 0.90 & 5.79 & 4.67 & 0.90 & \color{blue} \bfseries 4.68 \\
 & DPSM~\cite{shi2025direct}\textcolor{gray}{$_\text{ ICML'25}$} & 3.50 & 0.90 & 3.28 & \color{blue} \bfseries 3.09 & 0.90 & 3.45 & 4.08 & 0.90 & 3.49 & 5.54 & 0.90 & 5.83 & 4.39 & 0.90 & 5.19 \\
 & \cellcolor{Prune!20}\ours + HR (Ours) & \cellcolor{Prune!20}3.75 & \cellcolor{Prune!20}0.90 & \cellcolor{Prune!20}3.31 & \cellcolor{Prune!20}4.19 & \cellcolor{Prune!20}0.90 & \cellcolor{Prune!20}3.63 & \cellcolor{Prune!20}4.19 & \cellcolor{Prune!20}0.90 & \cellcolor{Prune!20}3.78 & \cellcolor{Prune!20}5.83 & \cellcolor{Prune!20}0.90 & \cellcolor{Prune!20}6.86 & \cellcolor{Prune!20}4.57 & \cellcolor{Prune!20}0.90 & \cellcolor{Prune!20}5.07 \\
 & \cellcolor{Prune!20}\ours + ALM (Ours) & \color{red} \bfseries \cellcolor{Prune!20}3.09 & \cellcolor{Prune!20}0.90 & \color{blue} \bfseries \cellcolor{Prune!20}3.10 & \color{red} \bfseries \cellcolor{Prune!20}3.06 & \cellcolor{Prune!20}0.90 & \color{red} \bfseries \cellcolor{Prune!20}3.21 & \color{red} \bfseries \cellcolor{Prune!20}3.22 & \cellcolor{Prune!20}0.90 & \color{blue} \bfseries \cellcolor{Prune!20}3.18 & \color{blue} \bfseries \cellcolor{Prune!20}5.50 & \cellcolor{Prune!20}0.90 & \color{red} \bfseries \cellcolor{Prune!20}5.10 & \color{red} \bfseries \cellcolor{Prune!20}3.83 & \cellcolor{Prune!20}0.90 & \color{red} \bfseries \cellcolor{Prune!20}4.60 \\
\bottomrule
\end{tabular}
}
\caption{\textbf{Extension of~\Cref{tab:results_gamma0.1} -- Results across image and text classification datasets (with RAPS \cite{angelopoulos2022raps} as non-conformity score).} Prediction set size (S), empirical coverage (C) and coverage gap (CG) values
with $\alpha=0.1$. The best model 
yields the smallest set size and coverage gap, while maintaining coverage close to the desired level. Best result in {\textbf{\color{red}Red}}, whereas {\textbf{\color{blue}Blue}} indicates the second best result.} 
\label{tab:results_gamma0.1_appendix}
\end{table*}

\paragraph{Varying imbalance factor $\gamma$ (extension of~\Cref{tab:ablation_gamma_cifar100lt_main}).}
\Cref{tab:ablation_gamma_cifar100lt_aps_raps} presents the performance when varying the imbalance severity of the training set using $\gamma$ (\ie, as $\gamma \to 1.0$, the per-class counts become equal). In line with the THR results in~\Cref{tab:ablation_gamma_cifar100lt_main}, \ours consistently outperforms all other existing conformal training objectives, as well as CE and FL, in terms of prediction set size and coverage gap. These findings remain consistent under different levels of training-set imbalance, further demonstrating the robustness of \ours{} to changes in class distribution.

\begin{table}[!htbp]
\centering
\resizebox{\linewidth}{!}{
\begin{tabular}{clcccccccccccc}
\toprule
& \multirow{2}{*}{Method} & \multicolumn{3}{c}{$\gamma=0.5$} & \multicolumn{3}{c}{$\gamma=0.8$} & \multicolumn{3}{c}{$\gamma=1.0$} \\
\cmidrule(lr){3-5} \cmidrule(lr){6-8} \cmidrule(lr){9-11} \cmidrule(lr){12-14}
 &  & S & C & CG & S & C & CG & S & C & CG \\
\midrule
\multirow[l]{8}{*}{\rotatebox[origin=c]{90}{APS}} & CE & 6.47 & 0.90 & 6.41 & 6.01 & 0.90 & 5.63 & 5.50 & 0.90 & 4.09 \\
 & FL~\cite{lin2018focalloss} & 8.79 & 0.90 & 5.61 & 7.07 & 0.90 & 4.78 & 5.49 & 0.90 & 4.62 \\
 & ConfTr~\cite{stutz2022conftr}\textcolor{gray}{$_\text{ ICLR'22}$} & 6.85 & 0.90 & 5.24 & 5.95 & 0.90 & 4.51 & \color{blue} \bfseries 4.57 & 0.90 & 3.82 \\
 & CUT~\cite{einbinder2022cut}\textcolor{gray}{$_\text{ NeurIPS'22}$} & \color{blue} \bfseries 5.09 & 0.90 & 4.00 & 4.90 & 0.90 & 3.99 & 4.87 & 0.90 & 3.80 \\
 & InfoCTr Fano~\cite{correia2025infocp}\textcolor{gray}{$_\text{ NeurIPS'24}$} & 6.16 & 0.90 & 4.22 & 5.66 & 0.90 & 3.93 & 4.94 & 0.90 & 3.84 \\
 & DPSM~\cite{shi2025direct}\textcolor{gray}{$_\text{ ICML'25}$} & 5.54 & 0.90 & \color{red} \bfseries 3.82 & 5.00 & 0.90 & 3.78 & 4.77 & 0.90 & 3.44 \\
 & \cellcolor{Prune!20}\ours + HR (Ours) & \cellcolor{Prune!20}5.49 & \cellcolor{Prune!20}0.90 & \cellcolor{Prune!20}4.09 & \color{blue} \bfseries \cellcolor{Prune!20}4.88 & \cellcolor{Prune!20}0.90 & \color{blue} \bfseries \cellcolor{Prune!20}3.76 & \cellcolor{Prune!20}4.76 & \cellcolor{Prune!20}0.90 & \color{blue} \bfseries \cellcolor{Prune!20}3.40 \\
 & \cellcolor{Prune!20}\ours + ALM (Ours) & \color{red} \bfseries \cellcolor{Prune!20}4.21 & \cellcolor{Prune!20}0.90 & \color{blue} \bfseries \cellcolor{Prune!20}3.89 & \color{red} \bfseries \cellcolor{Prune!20}4.02 & \cellcolor{Prune!20}0.90 & \color{red} \bfseries \cellcolor{Prune!20}3.70 & \color{red} \bfseries \cellcolor{Prune!20}3.99 & \cellcolor{Prune!20}0.90 & \color{red} \bfseries \cellcolor{Prune!20}3.22 \\
\midrule
\multirow[l]{8}{*}{\rotatebox[origin=c]{90}{RAPS}} & CE & 7.44 & 0.90 & 5.56 & 5.90 & 0.90 & 4.78 & 4.62 & 0.90 & 3.37 \\
 & FL~\cite{lin2018focalloss} & 8.97 & 0.90 & 6.08 & 5.75 & 0.90 & 4.81 & 4.68 & 0.90 & 3.63 \\
 & ConfTr~\cite{stutz2022conftr}\textcolor{gray}{$_\text{ ICLR'22}$} & 7.74 & 0.90 & 4.98 & 5.05 & 0.90 & 4.74 & 4.30 & 0.90 & 3.83 \\
 & CUT~\cite{einbinder2022cut}\textcolor{gray}{$_\text{ NeurIPS'22}$} & \color{blue} \bfseries 6.11 & 0.90 & 4.80 & 4.80 & 0.90 & \color{blue} \bfseries 4.01 & 4.55 & 0.90 & \color{red} \bfseries 3.16 \\
 & InfoCTr Fano~\cite{correia2025infocp}\textcolor{gray}{$_\text{ NeurIPS'24}$} & 6.13 & 0.90 & \color{blue} \bfseries 4.11 & \color{blue} \bfseries 4.66 & 0.90 & 4.02 & \color{blue} \bfseries 3.45 & 0.90 & 3.41 \\
 & DPSM~\cite{shi2025direct}\textcolor{gray}{$_\text{ ICML'25}$} & 7.02 & 0.90 & 4.61 & 4.74 & 0.90 & 4.51 & 4.08 & 0.90 & 3.49 \\
 & \cellcolor{Prune!20}\ours + HR (Ours) & \cellcolor{Prune!20}6.50 & \cellcolor{Prune!20}0.90 & \cellcolor{Prune!20}4.95 & \cellcolor{Prune!20}5.98 & \cellcolor{Prune!20}0.90 & \cellcolor{Prune!20}4.88 & \cellcolor{Prune!20}4.19 & \cellcolor{Prune!20}0.90 & \cellcolor{Prune!20}3.78 \\
 & \cellcolor{Prune!20}\ours + ALM (Ours) & \color{red} \bfseries \cellcolor{Prune!20}5.25 & \cellcolor{Prune!20}0.90 & \color{red} \bfseries \cellcolor{Prune!20}3.67 & \color{red} \bfseries \cellcolor{Prune!20}4.35 & \cellcolor{Prune!20}0.90 & \color{red} \bfseries \cellcolor{Prune!20}3.44 & \color{red} \bfseries \cellcolor{Prune!20}3.22 & \cellcolor{Prune!20}0.90 & \color{blue} \bfseries \cellcolor{Prune!20}3.18 \\
\bottomrule
\end{tabular}
}
\caption{\textbf{Extension of~\Cref{tab:ablation_gamma_cifar100lt_main} -- Ablation study of imbalance factor $\gamma$} on CIFAR100-LT with APS~\cite{romano2020aps} and RAPS~\cite{angelopoulos2022raps} as non-conformity scores. Performance comparison across different levels of class imbalance severity (lower values indicate more severe imbalance). Lower is better. \textbf{\color{Red}Red}: best result; \textbf{\color{Blue}Blue}: second best.}
\label{tab:ablation_gamma_cifar100lt_aps_raps}
\end{table}

\paragraph{Varying mis-coverage level $\alpha$ at test-time (extension of~\Cref{tab:ablation_alpha_cifar100lt}).}
\Cref{tab:ablation_alpha_cifar100lt_appendix} extends the results reported in \Cref{tab:ablation_alpha_cifar100lt} to APS and RAPS. The experiments were conducted on CIFAR100-LT with imbalance factor $\gamma = 0.1$. During training, we fixed the mis-coverage level to $\alpha = 0.01$. At test time, we varied the value of $\alpha$ in Split CP in order to examine whether the model overfits the training mis-coverage level. The observations made for THR also hold for APS and RAPS. In particular, the results show that \ours with ALM achieves the smallest set sizes and the lowest coverage gaps, and it performs better than the CE and FL objectives that do not depend on this parameter.

\begin{table}[!htbp]
\centering
\resizebox{\linewidth}{!}{
\begin{tabular}{clcccccccccccc}
\toprule
& \multirow{2}{*}{Method} & \multicolumn{3}{c}{$\alpha=0.01$} & \multicolumn{3}{c}{$\alpha=0.05$} & \multicolumn{3}{c}{$\alpha=0.1$} \\
\cmidrule(lr){3-5} \cmidrule(lr){6-8} \cmidrule(lr){9-11} \cmidrule(lr){12-14}
 &  & S & C & CG & S & C & CG & S & C & CG \\
\midrule
\multirow[l]{8}{*}{\rotatebox[origin=c]{90}{APS}} & CE & 21.75 & 0.99 & 4.96 & 13.34 & 0.95 & 5.26 & 9.20 & 0.90 & 6.80 \\
 & FL~\cite{lin2018focalloss} & 20.65 & 0.99 & 4.77 & 14.09 & 0.95 & 5.07 & 9.85 & 0.90 & 6.07 \\
 & ConfTr~\cite{stutz2022conftr}\textcolor{gray}{$_\text{ ICLR'22}$} & \color{blue} \bfseries 17.17 & 0.99 & 3.91 & 12.26 & 0.95 & 4.46 & 8.17 & 0.90 & 5.31 \\
 & CUT~\cite{einbinder2022cut}\textcolor{gray}{$_\text{ NeurIPS'22}$} & 18.69 & 0.99 & \color{blue} \bfseries 3.26 & 11.09 & 0.95 & \color{blue} \bfseries 3.69 & 6.97 & 0.90 & 4.55 \\
 & InfoCTr Fano~\cite{correia2025infocp}\textcolor{gray}{$_\text{ NeurIPS'24}$} & 19.53 & 0.99 & 3.65 & 10.95 & 0.95 & 4.27 & 7.18 & 0.90 & 5.65 \\
 & DPSM~\cite{shi2025direct}\textcolor{gray}{$_\text{ ICML'25}$} & 18.41 & 0.99 & 3.73 & 10.82 & 0.95 & 4.10 & 6.60 & 0.90 & 4.76 \\
 & \cellcolor{Prune!20}\ours + HR (Ours) & \cellcolor{Prune!20}17.34 & \cellcolor{Prune!20}0.99 & \cellcolor{Prune!20}3.53 & \color{blue} \bfseries \cellcolor{Prune!20}9.78 & \cellcolor{Prune!20}0.95 & \cellcolor{Prune!20}4.08 & \color{blue} \bfseries \cellcolor{Prune!20}5.65 & \cellcolor{Prune!20}0.90 & \color{blue} \bfseries \cellcolor{Prune!20}4.12 \\
 & \cellcolor{Prune!20}\ours + ALM (Ours) & \color{red} \bfseries \cellcolor{Prune!20}16.07 & \cellcolor{Prune!20}0.99 & \color{red} \bfseries \cellcolor{Prune!20}3.07 & \color{red} \bfseries \cellcolor{Prune!20}8.62 & \cellcolor{Prune!20}0.95 & \color{red} \bfseries \cellcolor{Prune!20}3.66 & \color{red} \bfseries \cellcolor{Prune!20}4.49 & \cellcolor{Prune!20}0.90 & \color{red} \bfseries \cellcolor{Prune!20}4.10 \\
\midrule
\multirow[l]{8}{*}{\rotatebox[origin=c]{90}{RAPS}} & CE & 18.19 & 0.99 & 4.55 & 13.23 & 0.95 & 4.74 & 9.15 & 0.90 & 6.12 \\
 & FL~\cite{lin2018focalloss} & 19.11 & 0.99 & 3.43 & 13.94 & 0.95 & 4.94 & 9.85 & 0.90 & 6.60 \\
 & ConfTr~\cite{stutz2022conftr}\textcolor{gray}{$_\text{ ICLR'22}$} & 20.95 & 0.99 & 3.50 & 13.51 & 0.95 & \color{blue} \bfseries 3.79 & 9.23 & 0.90 & 5.16 \\
 & CUT~\cite{einbinder2022cut}\textcolor{gray}{$_\text{ NeurIPS'22}$} & \color{blue} \bfseries 16.21 & 0.99 & 3.53 & 12.03 & 0.95 & 3.85 & 8.03 & 0.90 & 4.90 \\
 & InfoCTr Fano~\cite{correia2025infocp}\textcolor{gray}{$_\text{ NeurIPS'24}$} & 16.62 & 0.99 & \color{blue} \bfseries 3.23 & 12.55 & 0.95 & 3.95 & 7.21 & 0.90 & \color{blue} \bfseries 4.29 \\
 & DPSM~\cite{shi2025direct}\textcolor{gray}{$_\text{ ICML'25}$} & 17.17 & 0.99 & \color{red} \bfseries 3.18 & 11.98 & 0.95 & 3.96 & \color{blue} \bfseries 7.18 & 0.90 & 4.87 \\
 & \cellcolor{Prune!20}\ours + HR (Ours) & \cellcolor{Prune!20}17.14 & \cellcolor{Prune!20}0.99 & \cellcolor{Prune!20}4.02 & \color{blue} \bfseries \cellcolor{Prune!20}11.27 & \cellcolor{Prune!20}0.95 & \cellcolor{Prune!20}4.19 & \cellcolor{Prune!20}7.53 & \cellcolor{Prune!20}0.90 & \cellcolor{Prune!20}5.06 \\
 & \cellcolor{Prune!20}\ours + ALM (Ours) & \color{red} \bfseries \cellcolor{Prune!20}15.89 & \cellcolor{Prune!20}0.99 & \cellcolor{Prune!20}3.38 & \color{red} \bfseries \cellcolor{Prune!20}10.34 & \cellcolor{Prune!20}0.95 & \color{red} \bfseries \cellcolor{Prune!20}3.58 & \color{red} \bfseries \cellcolor{Prune!20}6.22 & \cellcolor{Prune!20}0.90 & \color{red} \bfseries \cellcolor{Prune!20}4.01 \\
\bottomrule
\end{tabular}
}
\caption{\textbf{Extension of~\Cref{tab:ablation_alpha_cifar100lt} -- Ablation study of test-time mis-coverage level $\alpha$} on CIFAR100-LT (the imbalance factor $\gamma$ is set $0.1$) with APS~\cite{romano2020aps} and RAPS~\cite{angelopoulos2022raps} as non-conformity scores. Lower is better. \textbf{\color{Red}Red}: best result; \textbf{\color{Blue}Blue}: second best.}
\label{tab:ablation_alpha_cifar100lt_appendix}
\end{table}

\begin{figure*}[!htbp]
    \centering
    \begin{subfigure}[b]{\textwidth}
        \centering
        \includegraphics[width=\textwidth]{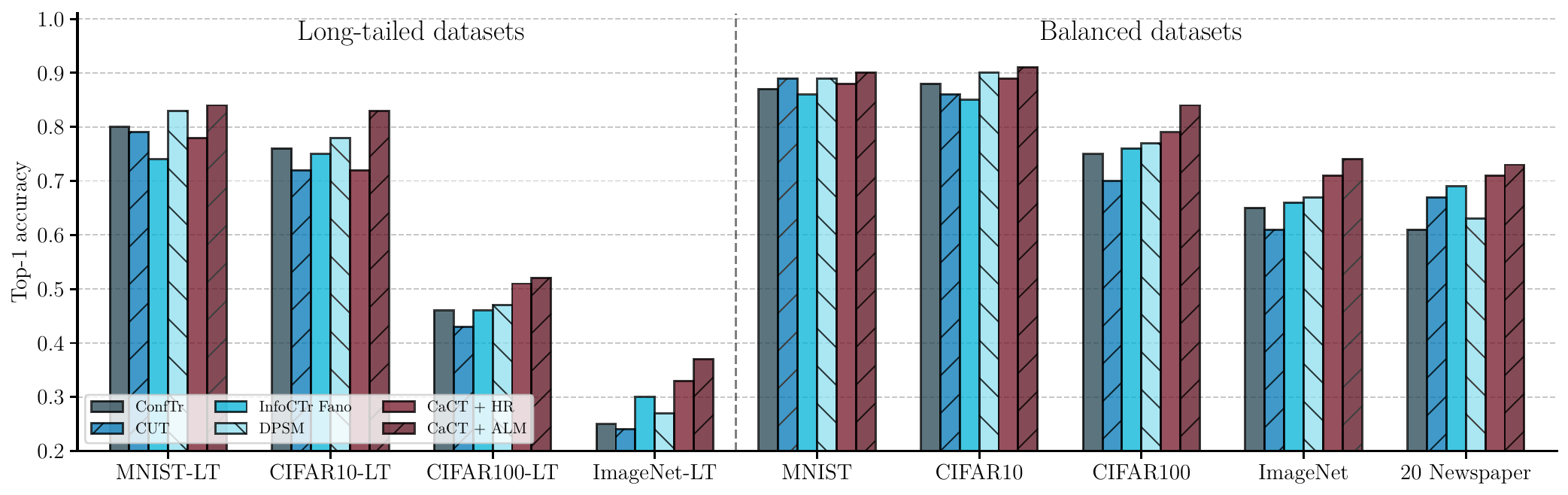}
        \caption{Top-1 accuracy.}
        \label{fig:top1}
    \end{subfigure}
    
    \vspace{0.0cm}
    
    \begin{subfigure}[b]{\textwidth}
        \centering
        \includegraphics[width=\textwidth]{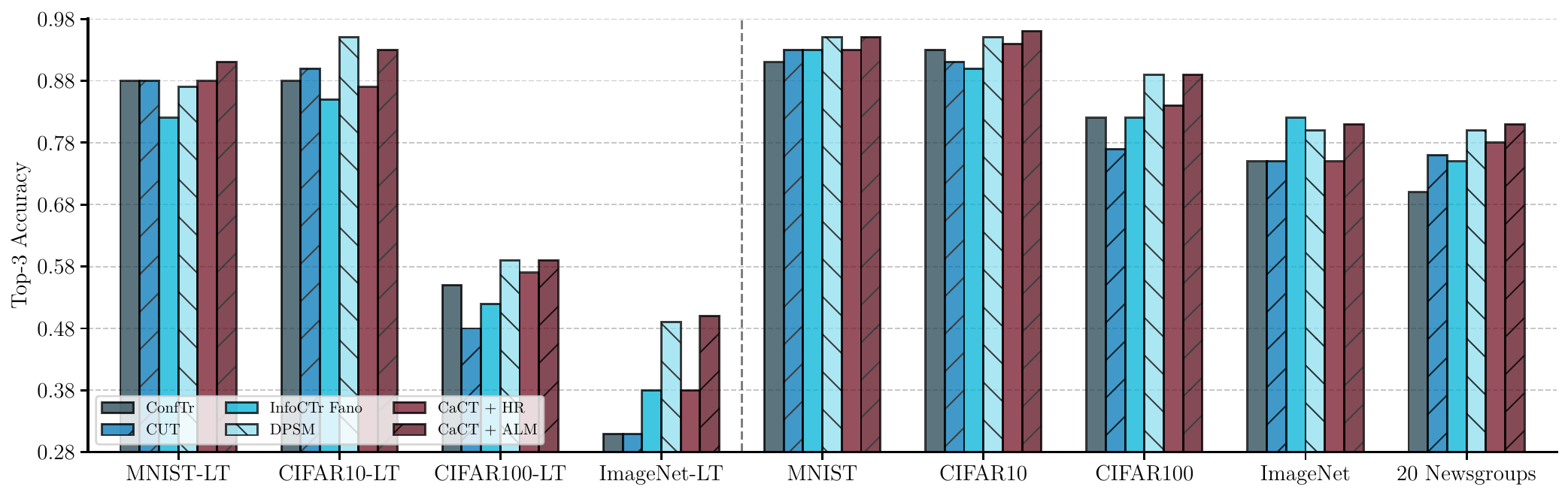}
        \caption{Top-3 accuracy.}
        \label{fig:top3}
    \end{subfigure}
    \caption{\textbf{Comparison of the Top-$k$ ($k=1, 3$) performance of different conformal training objectives} across long-tailed and balanced image and text classification datasets.}
    \label{fig:top13}
\end{figure*}

\paragraph{Top-$k$ $(k={1,3})$ performance (extension of~\Cref{fig:topk_main}).}
\Cref{fig:top13} extends the results in~\Cref{fig:topk_main} to other long-tailed and balanced image and text classification datasets. \ours achieves comparable or better top-1 and top-3 accuracies compared to other baselines. Combined with the superior performance of \ours in terms of prediction set size and coverage gap reported in~\Cref{tab:results_gamma0.1_main,tab:results_gamma0.1,tab:results_gamma0.1_appendix_LT,tab:results_gamma0.1_appendix}, these results demonstrate that \ours effectively balances model accuracy with the predictive efficiency of CP.




\paragraph{Beyond Split CP.}
All experiments have been conducted under the Split CP setting. We further evaluate whether the same findings hold across different settings, specifically LabelCP, which is a Mondrian conformal variant~\cite{vovk2005book} and ClusterCP~\cite{ding2023clustercp}. Details on how prediction sets are produced with LabelCP and ClusterCP are provided in~\Cref{sec:Conditional Conformal Prediction}. To complement the results in main paper, \Cref{tab:labelcp_gamma0.1,tab:clustercp_gamma0.1} present the performance of all methods on image and text classification datasets under these two CP algorithms.
The results show that \ours consistently outperforms the other objectives, regardless of the CP procedure or the choice of non conformity score. This superiority comes from the fact that \ours, combined with ALM, allows the model to adapt its behavior in a class-conditionally fashion, which leads to more reliable and better shaped prediction set sizes.

\begin{table*}[!htbp]
\centering
\resizebox{\textwidth}{!}{
\begin{tabular}{cclccccccccccccccc}
\toprule
& \multirow{2}{*}{Score} & \multirow{2}{*}{Method} & \multicolumn{3}{c}{MNIST-LT} & \multicolumn{3}{c}{CIFAR10-LT} & \multicolumn{3}{c}{CIFAR100-LT} & \multicolumn{3}{c}{ImageNet-LT} & \multicolumn{3}{c}{20 Newsgroups} \\
\cmidrule(lr){4-6} \cmidrule(lr){7-9} \cmidrule(lr){10-12} \cmidrule(lr){13-15} \cmidrule(lr){16-18}
& & & S & C & CG & S & C & CG & S & C & CG & S & C & CG & S & C & CG\\
\cmidrule(lr){2-18}
\multirow[l]{24}{*}{\rotatebox[origin=c]{90}{Label CP}} & \multirow[l]{8}{*}{THR} & CE & 7.21 & 0.90 & 5.33 & 7.70 & 0.90 & 5.76 & 6.00 & 0.90 & 6.75 & 6.45 & 0.90 & 5.01 & 6.60 & 0.90 & 7.54 \\
 &  & FL~\cite{lin2018focalloss} & 5.61 & 0.90 & 6.03 & 4.86 & 0.90 & 7.46 & 5.48 & 0.90 & 7.46 & 6.09 & 0.90 & 6.88 & 4.55 & 0.90 & 7.06 \\
 &  & ConfTr~\cite{stutz2022conftr}\textcolor{gray}{$_\text{ ICLR'22}$} & 5.42 & 0.90 & 6.11 & 4.52 & 0.90 & 7.66 & 5.42 & 0.90 & 7.55 & 6.05 & 0.90 & 7.10 & 4.30 & 0.90 & 7.00 \\
 &  & CUT~\cite{einbinder2022cut}\textcolor{gray}{$_\text{ NeurIPS'22}$} & 4.97 & 0.90 & 5.81 & 4.42 & 0.90 & 7.19 & 5.06 & 0.90 & 7.29 & 5.59 & 0.90 & \color{blue} \bfseries 4.45 & 4.23 & 0.90 & 6.83 \\
 &  & InfoCTr Fano~\cite{correia2025infocp}\textcolor{gray}{$_\text{ NeurIPS'24}$} & 4.15 & 0.90 & 7.70 & 5.01 & 0.90 & \color{blue} \bfseries 5.25 & 5.60 & 0.90 & 7.15 & 5.65 & 0.90 & 6.80 & 4.00 & 0.90 & 6.50 \\
 &  & DPSM~\cite{shi2025direct}\textcolor{gray}{$_\text{ ICML'25}$} & 3.57 & 0.90 & 5.27 & \color{red} \bfseries 3.73 & 0.90 & 5.26 & 3.91 & 0.90 & 6.70 & \color{red} \bfseries 4.99 & 0.90 & 5.57 & \color{red} \bfseries 3.81 & 0.90 & 6.88 \\
 &  & \cellcolor{Prune!20}\ours + HR (Ours) & \color{blue} \bfseries \cellcolor{Prune!20}3.47 & \cellcolor{Prune!20}0.90 & \color{blue} \bfseries \cellcolor{Prune!20}4.82 & \cellcolor{Prune!20}4.10 & \cellcolor{Prune!20}0.90 & \cellcolor{Prune!20}5.61 & \color{blue} \bfseries \cellcolor{Prune!20}3.87 & \cellcolor{Prune!20}0.90 & \color{blue} \bfseries \cellcolor{Prune!20}6.42 & \cellcolor{Prune!20}5.64 & \cellcolor{Prune!20}0.90 & \cellcolor{Prune!20}6.11 & \cellcolor{Prune!20}3.98 & \cellcolor{Prune!20}0.90 & \color{blue} \bfseries \cellcolor{Prune!20}6.27 \\
 &  & \cellcolor{Prune!20}\ours + ALM (Ours) & \color{red} \bfseries \cellcolor{Prune!20}2.75 & \cellcolor{Prune!20}0.90 & \color{red} \bfseries \cellcolor{Prune!20}4.35 & \color{blue} \bfseries \cellcolor{Prune!20}3.95 & \cellcolor{Prune!20}0.90 & \color{red} \bfseries \cellcolor{Prune!20}4.85 & \color{red} \bfseries \cellcolor{Prune!20}3.30 & \cellcolor{Prune!20}0.90 & \color{red} \bfseries \cellcolor{Prune!20}6.01 & \color{blue} \bfseries \cellcolor{Prune!20}5.25 & \cellcolor{Prune!20}0.90 & \color{red} \bfseries \cellcolor{Prune!20}4.38 & \color{blue} \bfseries \cellcolor{Prune!20}3.86 & \cellcolor{Prune!20}0.90 & \color{red} \bfseries \cellcolor{Prune!20}6.00 \\
\cmidrule(lr){2-18}
 & \multirow[l]{8}{*}{APS} & CE & 7.47 & 0.90 & \color{blue} \bfseries 3.44 & 5.00 & 0.90 & 4.97 & 8.50 & 0.90 & 8.10 & 8.65 & 0.90 & 5.09 & 6.73 & 0.90 & 7.02 \\
 &  & FL~\cite{lin2018focalloss} & 7.15 & 0.90 & 3.88 & 5.26 & 0.90 & 5.23 & 7.38 & 0.90 & 7.99 & 7.87 & 0.90 & 4.80 & 5.97 & 0.90 & 6.91 \\
 &  & ConfTr~\cite{stutz2022conftr}\textcolor{gray}{$_\text{ ICLR'22}$} & 6.50 & 0.90 & 4.76 & 5.78 & 0.90 & 5.75 & 5.15 & 0.90 & 7.76 & 6.30 & 0.90 & 4.23 & 4.46 & 0.90 & 6.70 \\
 &  & CUT~\cite{einbinder2022cut}\textcolor{gray}{$_\text{ NeurIPS'22}$} & 5.53 & 0.90 & 4.37 & 5.24 & 0.90 & 5.31 & 4.90 & 0.90 & 7.15 & \color{blue} \bfseries 5.74 & 0.90 & 4.06 & 4.26 & 0.90 & 6.44 \\
 &  & InfoCTr Fano~\cite{correia2025infocp}\textcolor{gray}{$_\text{ NeurIPS'24}$} & 4.25 & 0.90 & 5.09 & 5.15 & 0.90 & 5.40 & 4.80 & 0.90 & 7.32 & 5.95 & 0.90 & 3.82 & 4.11 & 0.90 & 6.30 \\
 &  & DPSM~\cite{shi2025direct}\textcolor{gray}{$_\text{ ICML'25}$} & 4.55 & 0.90 & 3.83 & 4.97 & 0.90 & 4.67 & 5.03 & 0.90 & 6.71 & 5.86 & 0.90 & 4.21 & 4.14 & 0.90 & \color{red} \bfseries 5.35 \\
 &  & \cellcolor{Prune!20}\ours + HR (Ours) & \color{blue} \bfseries \cellcolor{Prune!20}3.86 & \cellcolor{Prune!20}0.90 & \cellcolor{Prune!20}3.71 & \color{blue} \bfseries \cellcolor{Prune!20}4.32 & \cellcolor{Prune!20}0.90 & \color{blue} \bfseries \cellcolor{Prune!20}4.56 & \color{blue} \bfseries \cellcolor{Prune!20}4.46 & \cellcolor{Prune!20}0.90 & \color{blue} \bfseries \cellcolor{Prune!20}6.11 & \cellcolor{Prune!20}5.79 & \cellcolor{Prune!20}0.90 & \color{blue} \bfseries \cellcolor{Prune!20}3.78 & \color{blue} \bfseries \cellcolor{Prune!20}3.91 & \cellcolor{Prune!20}0.90 & \cellcolor{Prune!20}6.00 \\
 &  & \cellcolor{Prune!20}\ours + ALM (Ours) & \color{red} \bfseries \cellcolor{Prune!20}3.09 & \cellcolor{Prune!20}0.90 & \color{red} \bfseries \cellcolor{Prune!20}3.40 & \color{red} \bfseries \cellcolor{Prune!20}3.90 & \cellcolor{Prune!20}0.90 & \color{red} \bfseries \cellcolor{Prune!20}4.21 & \color{red} \bfseries \cellcolor{Prune!20}4.26 & \cellcolor{Prune!20}0.90 & \color{red} \bfseries \cellcolor{Prune!20}5.63 & \color{red} \bfseries \cellcolor{Prune!20}5.35 & \cellcolor{Prune!20}0.90 & \color{red} \bfseries \cellcolor{Prune!20}3.65 & \color{red} \bfseries \cellcolor{Prune!20}3.75 & \cellcolor{Prune!20}0.90 & \color{blue} \bfseries \cellcolor{Prune!20}5.80 \\
\cmidrule(lr){2-18}
 & \multirow[l]{8}{*}{RAPS} & CE & 6.86 & 0.90 & 3.45 & \color{blue} \bfseries 4.98 & 0.90 & 4.38 & 5.05 & 0.90 & 6.17 & 6.70 & 0.90 & 4.62 & 5.85 & 0.90 & 5.45 \\
 &  & FL~\cite{lin2018focalloss} & 6.58 & 0.90 & 4.31 & 5.51 & 0.90 & 5.42 & 4.51 & 0.90 & 7.22 & 5.67 & 0.90 & 4.41 & 4.89 & 0.90 & 6.26 \\
 &  & ConfTr~\cite{stutz2022conftr}\textcolor{gray}{$_\text{ ICLR'22}$} & 6.48 & 0.90 & 4.62 & 5.70 & 0.90 & 5.79 & 4.32 & 0.90 & 7.60 & \color{blue} \bfseries 5.30 & 0.90 & 4.33 & 4.55 & 0.90 & 6.55 \\
 &  & CUT~\cite{einbinder2022cut}\textcolor{gray}{$_\text{ NeurIPS'22}$} & 5.96 & 0.90 & \color{blue} \bfseries 3.44 & 5.52 & 0.90 & \color{blue} \bfseries 4.24 & 4.22 & 0.90 & \color{blue} \bfseries 5.23 & \color{red} \bfseries 5.06 & 0.90 & 4.47 & 4.49 & 0.90 & \color{blue} \bfseries 5.33 \\
 &  & InfoCTr Fano~\cite{correia2025infocp}\textcolor{gray}{$_\text{ NeurIPS'24}$} & \color{blue} \bfseries 4.21 & 0.90 & 3.75 & 5.30 & 0.90 & 5.30 & 4.10 & 0.90 & 7.14 & 5.70 & 0.90 & 4.05 & \color{blue} \bfseries 4.20 & 0.90 & 6.15 \\
 &  & DPSM~\cite{shi2025direct}\textcolor{gray}{$_\text{ ICML'25}$} & 4.47 & 0.90 & 3.46 & 5.73 & 0.90 & 4.36 & \color{blue} \bfseries 3.72 & 0.90 & 6.63 & 6.07 & 0.90 & 4.09 & 4.23 & 0.90 & 6.25 \\
 &  & \cellcolor{Prune!20}\ours + HR (Ours) & \cellcolor{Prune!20}4.90 & \cellcolor{Prune!20}0.90 & \cellcolor{Prune!20}4.08 & \cellcolor{Prune!20}5.15 & \cellcolor{Prune!20}0.90 & \cellcolor{Prune!20}4.87 & \cellcolor{Prune!20}4.03 & \cellcolor{Prune!20}0.90 & \cellcolor{Prune!20}6.47 & \cellcolor{Prune!20}5.58 & \cellcolor{Prune!20}0.90 & \color{blue} \bfseries \cellcolor{Prune!20}3.86 & \cellcolor{Prune!20}4.38 & \cellcolor{Prune!20}0.90 & \cellcolor{Prune!20}5.86 \\
 &  & \cellcolor{Prune!20}\ours + ALM (Ours) & \color{red} \bfseries \cellcolor{Prune!20}2.98 & \cellcolor{Prune!20}0.90 & \color{red} \bfseries \cellcolor{Prune!20}3.42 & \color{red} \bfseries \cellcolor{Prune!20}4.49 & \cellcolor{Prune!20}0.90 & \color{red} \bfseries \cellcolor{Prune!20}3.75 & \color{red} \bfseries \cellcolor{Prune!20}3.67 & \cellcolor{Prune!20}0.90 & \color{red} \bfseries \cellcolor{Prune!20}5.09 & \cellcolor{Prune!20}5.70 & \cellcolor{Prune!20}0.90 & \color{red} \bfseries \cellcolor{Prune!20}3.29 & \color{red} \bfseries \cellcolor{Prune!20}4.17 & \cellcolor{Prune!20}0.90 & \color{red} \bfseries \cellcolor{Prune!20}5.03 \\
\bottomrule
\end{tabular}
}
\caption{\textbf{Results across long-tailed image classification datasets using LabelCP.} Prediction set size (S), empirical coverage (C) and coverage gap (CG) values
with $\alpha=0.1$ and an imbalance factor of $\gamma=0.1$ for all datasets except ImageNet-LT~\cite{liu2019imagenetlt}. The best model 
yields the smallest set size and coverage gap, while maintaining coverage close to the desired level. Best result in {\textbf{\color{red}Red}}, whereas {\textbf{\color{blue}Blue}} indicates the second best result.} 
\label{tab:labelcp_gamma0.1}
\end{table*}
\begin{table*}[!htbp]
\centering
\scriptsize
\resizebox{\textwidth}{!}{
\begin{tabular}{cclccccccccccccccc}
\toprule
& \multirow{2}{*}{Score} & \multirow{2}{*}{Method} & \multicolumn{3}{c}{MNIST-LT} & \multicolumn{3}{c}{CIFAR10-LT} & \multicolumn{3}{c}{CIFAR100-LT} & \multicolumn{3}{c}{ImageNet-LT}  & \multicolumn{3}{c}{20 Newsgroups}\\
\cmidrule(lr){4-6} \cmidrule(lr){7-9} \cmidrule(lr){10-12} \cmidrule(lr){13-15} \cmidrule(lr){16-18}
& & & S & C & CG & S & C & CG & S & C & CG & S & C & CG & S & C & CG\\
\cmidrule(lr){2-18}
\multirow[l]{24}{*}{\rotatebox[origin=c]{90}{Cluster CP}} & \multirow[l]{8}{*}{THR} & CE & 6.65 & 0.90 & 4.83 & 6.95 & 0.90 & \color{blue} \bfseries 4.99 & 5.65 & 0.90 & 5.98 & 8.45 & 0.90 & 4.40 & 6.50 & 0.90 & 6.95 \\
 &  & FL~\cite{lin2018focalloss} & 6.44 & 0.90 & 4.56 & 6.08 & 0.90 & 5.60 & 4.41 & 0.90 & 4.88 & 6.43 & 0.90 & 4.02 & 5.89 & 0.90 & 6.98 \\
 &  & ConfTr~\cite{stutz2022conftr}\textcolor{gray}{$_\text{ ICLR'22}$} & 5.10 & 0.90 & 4.44 & 5.50 & 0.90 & 5.50 & 3.87 & 0.90 & 4.10 & 5.63 & 0.90 & 3.42 & 4.68 & 0.90 & 6.58 \\
 &  & CUT~\cite{einbinder2022cut}\textcolor{gray}{$_\text{ NeurIPS'22}$} & 4.44 & 0.90 & 4.32 & 4.51 & 0.90 & 5.17 & 3.96 & 0.90 & 3.50 & 5.31 & 0.90 & 3.48 & \color{red} \bfseries 3.43 & 0.90 & 6.22 \\
 &  & InfoCTr Fano~\cite{correia2025infocp}\textcolor{gray}{$_\text{ NeurIPS'24}$} & 4.08 & 0.90 & 3.96 & 4.64 & 0.90 & 5.40 & 3.75 & 0.90 & 3.73 & 6.06 & 0.90 & 3.48 & 4.88 & 0.90 & 6.15 \\
 &  & DPSM~\cite{shi2025direct}\textcolor{gray}{$_\text{ ICML'25}$} & 4.46 & 0.90 & \color{red} \bfseries 3.51 & 4.59 & 0.90 & \color{red} \bfseries 4.78 & 3.71 & 0.90 & 3.68 & \color{blue} \bfseries 5.10 & 0.90 & \color{red} \bfseries 2.67 & 3.51 & 0.90 & \color{red} \bfseries 5.86 \\
 &  & \cellcolor{Prune!20}\ours + HR (Ours) & \color{blue} \bfseries \cellcolor{Prune!20}3.85 & \cellcolor{Prune!20}0.90 & \cellcolor{Prune!20}3.84 & \color{blue} \bfseries \cellcolor{Prune!20}4.29 & \cellcolor{Prune!20}0.90 & \cellcolor{Prune!20}5.03 & \color{blue} \bfseries \cellcolor{Prune!20}3.42 & \cellcolor{Prune!20}0.90 & \color{red} \bfseries \cellcolor{Prune!20}3.19 & \color{red} \bfseries \cellcolor{Prune!20}5.05 & \cellcolor{Prune!20}0.90 & \color{blue} \bfseries \cellcolor{Prune!20}3.35 & \cellcolor{Prune!20}3.86 & \cellcolor{Prune!20}0.90 & \cellcolor{Prune!20}5.98 \\
 &  & \cellcolor{Prune!20}\ours + ALM (Ours) & \color{red} \bfseries \cellcolor{Prune!20}2.63 & \cellcolor{Prune!20}0.90 & \color{blue} \bfseries \cellcolor{Prune!20}3.56 & \color{red} \bfseries \cellcolor{Prune!20}3.47 & \cellcolor{Prune!20}0.90 & \cellcolor{Prune!20}5.15 & \color{red} \bfseries \cellcolor{Prune!20}3.16 & \cellcolor{Prune!20}0.90 & \color{blue} \bfseries \cellcolor{Prune!20}3.42 & \cellcolor{Prune!20}5.22 & \cellcolor{Prune!20}0.90 & \cellcolor{Prune!20}3.60 & \color{blue} \bfseries \cellcolor{Prune!20}3.47 & \cellcolor{Prune!20}0.90 & \color{blue} \bfseries \cellcolor{Prune!20}5.88 \\
\cmidrule(lr){2-18}
 & \multirow[l]{8}{*}{APS} & CE & 6.70 & 0.90 & 4.40 & 6.11 & 0.90 & \color{blue} \bfseries 4.06 & 9.13 & 0.90 & 6.67 & 7.98 & 0.90 & 4.51 & 7.08 & 0.90 & 6.10 \\
 &  & FL~\cite{lin2018focalloss} & 6.47 & 0.90 & 4.24 & 5.48 & 0.90 & 5.05 & 7.65 & 0.90 & 6.30 & 6.12 & 0.90 & 4.18 & 5.72 & 0.90 & 5.96 \\
 &  & ConfTr~\cite{stutz2022conftr}\textcolor{gray}{$_\text{ ICLR'22}$} & 6.07 & 0.90 & 3.79 & 5.22 & 0.90 & 4.66 & 6.01 & 0.90 & 6.35 & 5.76 & 0.90 & 4.00 & 5.31 & 0.90 & 5.88 \\
 &  & CUT~\cite{einbinder2022cut}\textcolor{gray}{$_\text{ NeurIPS'22}$} & 4.78 & 0.90 & \color{red} \bfseries 3.47 & 4.42 & 0.90 & 4.20 & \color{blue} \bfseries 4.01 & 0.90 & 5.90 & \color{red} \bfseries 5.49 & 0.90 & 3.57 & \color{blue} \bfseries 4.68 & 0.90 & 5.20 \\
 &  & InfoCTr Fano~\cite{correia2025infocp}\textcolor{gray}{$_\text{ NeurIPS'24}$} & 4.57 & 0.90 & 3.97 & 3.73 & 0.90 & 5.29 & 6.05 & 0.90 & 6.08 & 5.80 & 0.90 & 3.65 & 5.30 & 0.90 & 5.33 \\
 &  & DPSM~\cite{shi2025direct}\textcolor{gray}{$_\text{ ICML'25}$} & 4.72 & 0.90 & 3.65 & \color{red} \bfseries 3.12 & 0.90 & 5.11 & 4.03 & 0.90 & 5.49 & 5.61 & 0.90 & 3.85 & 5.52 & 0.90 & 5.02 \\
 &  & \cellcolor{Prune!20}\ours + HR (Ours) & \color{blue} \bfseries \cellcolor{Prune!20}4.55 & \cellcolor{Prune!20}0.90 & \cellcolor{Prune!20}3.97 & \cellcolor{Prune!20}3.55 & \cellcolor{Prune!20}0.90 & \cellcolor{Prune!20}5.87 & \cellcolor{Prune!20}4.54 & \cellcolor{Prune!20}0.90 & \color{blue} \bfseries \cellcolor{Prune!20}5.33 & \color{blue} \bfseries \cellcolor{Prune!20}5.60 & \cellcolor{Prune!20}0.90 & \color{blue} \bfseries \cellcolor{Prune!20}3.48 & \cellcolor{Prune!20}4.86 & \cellcolor{Prune!20}0.90 & \color{blue} \bfseries \cellcolor{Prune!20}4.90 \\
 &  & \cellcolor{Prune!20}\ours + ALM (Ours) & \color{red} \bfseries \cellcolor{Prune!20}4.31 & \cellcolor{Prune!20}0.90 & \color{blue} \bfseries \cellcolor{Prune!20}3.51 & \color{blue} \bfseries \cellcolor{Prune!20}3.26 & \cellcolor{Prune!20}0.90 & \color{red} \bfseries \cellcolor{Prune!20}3.58 & \color{red} \bfseries \cellcolor{Prune!20}3.88 & \cellcolor{Prune!20}0.90 & \color{red} \bfseries \cellcolor{Prune!20}4.83 & \cellcolor{Prune!20}5.87 & \cellcolor{Prune!20}0.90 & \color{red} \bfseries \cellcolor{Prune!20}3.29 & \color{red} \bfseries \cellcolor{Prune!20}4.09 & \cellcolor{Prune!20}0.90 & \color{red} \bfseries \cellcolor{Prune!20}4.81 \\
\cmidrule(lr){2-18}
 & \multirow[l]{8}{*}{RAPS} & CE & 6.65 & 0.90 & 4.10 & 6.10 & 0.90 & 6.83 & 5.99 & 0.90 & 6.10 & 6.07 & 0.90 & 5.19 & 6.67 & 0.90 & 5.24 \\
 &  & FL~\cite{lin2018focalloss} & 5.07 & 0.90 & 3.49 & 5.53 & 0.90 & 3.65 & 5.00 & 0.90 & 6.08 & 5.64 & 0.90 & 4.65 & 5.65 & 0.90 & 5.13 \\
 &  & ConfTr~\cite{stutz2022conftr}\textcolor{gray}{$_\text{ ICLR'22}$} & 4.29 & 0.90 & 3.44 & 4.38 & 0.90 & 3.19 & \color{red} \bfseries 3.92 & 0.90 & 5.34 & \color{red} \bfseries 5.32 & 0.90 & 4.17 & \color{blue} \bfseries 4.24 & 0.90 & 5.06 \\
 &  & CUT~\cite{einbinder2022cut}\textcolor{gray}{$_\text{ NeurIPS'22}$} & \color{blue} \bfseries 3.26 & 0.90 & \color{red} \bfseries 3.12 & 3.98 & 0.90 & 2.89 & \color{blue} \bfseries 4.52 & 0.90 & 5.39 & 5.80 & 0.90 & 3.68 & 4.46 & 0.90 & 4.66 \\
 &  & InfoCTr Fano~\cite{correia2025infocp}\textcolor{gray}{$_\text{ NeurIPS'24}$} & 3.73 & 0.90 & 3.60 & 4.31 & 0.90 & 3.36 & 5.42 & 0.90 & 5.19 & 5.91 & 0.90 & 3.60 & 4.82 & 0.90 & 4.84 \\
 &  & DPSM~\cite{shi2025direct}\textcolor{gray}{$_\text{ ICML'25}$} & 3.36 & 0.90 & \color{blue} \bfseries 3.33 & \color{blue} \bfseries 3.05 & 0.90 & 5.07 & 5.82 & 0.90 & \color{red} \bfseries 4.69 & 6.22 & 0.90 & 3.80 & 4.54 & 0.90 & \color{blue} \bfseries 4.09 \\
 &  & \cellcolor{Prune!20}\ours + HR (Ours) & \color{red} \bfseries \cellcolor{Prune!20}3.06 & \cellcolor{Prune!20}0.90 & \cellcolor{Prune!20}3.74 & \cellcolor{Prune!20}3.47 & \cellcolor{Prune!20}0.90 & \color{blue} \bfseries \cellcolor{Prune!20}2.64 & \cellcolor{Prune!20}5.90 & \cellcolor{Prune!20}0.90 & \color{blue} \bfseries \cellcolor{Prune!20}5.18 & \cellcolor{Prune!20}5.93 & \cellcolor{Prune!20}0.90 & \color{blue} \bfseries \cellcolor{Prune!20}3.45 & \cellcolor{Prune!20}4.26 & \cellcolor{Prune!20}0.90 & \cellcolor{Prune!20}4.52 \\
 &  & \cellcolor{Prune!20}\ours + ALM (Ours) & \cellcolor{Prune!20}3.43 & \cellcolor{Prune!20}0.90 & \cellcolor{Prune!20}3.36 & \color{red} \bfseries \cellcolor{Prune!20}2.95 & \cellcolor{Prune!20}0.90 & \color{red} \bfseries \cellcolor{Prune!20}2.48 & \cellcolor{Prune!20}4.70 & \cellcolor{Prune!20}0.90 & \cellcolor{Prune!20}5.29 & \color{blue} \bfseries \cellcolor{Prune!20}5.48 & \cellcolor{Prune!20}0.90 & \color{red} \bfseries \cellcolor{Prune!20}3.01 & \color{red} \bfseries \cellcolor{Prune!20}3.93 & \cellcolor{Prune!20}0.90 & \color{red} \bfseries \cellcolor{Prune!20}3.93 \\

\bottomrule
\end{tabular}
}
\caption{\textbf{Results across long-tailed image classification datasets using ClusterCP.} Prediction set size (S), empirical coverage (C) and coverage gap (CG) values
with $\alpha=0.1$ and an imbalance factor of $\gamma=0.1$ for all datasets except ImageNet-LT~\cite{liu2019imagenetlt}. The best model 
yields the smallest set size and coverage gap, while maintaining coverage close to the desired level. Best result in {\textbf{\color{red}Red}}, whereas {\textbf{\color{blue}Blue}} indicates the second best result.} 
\label{tab:clustercp_gamma0.1}
\end{table*}

\begin{figure}[!htbp]
    \centering
    \includegraphics[width=\columnwidth]{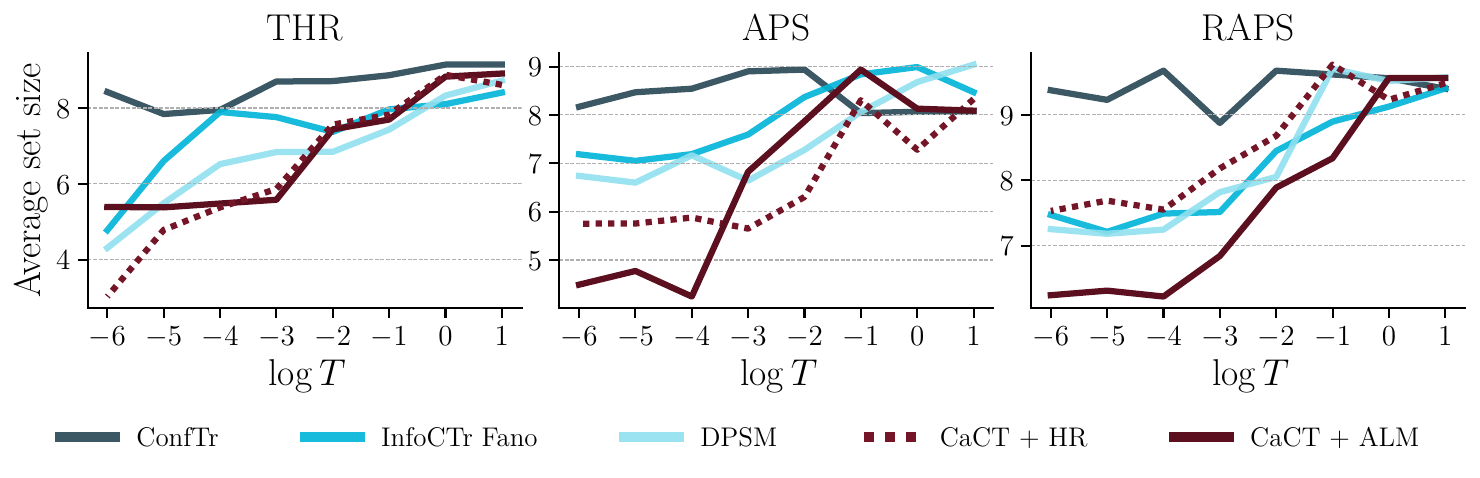}
    \caption{\textbf{Ablation study on the effect of temperature parameter $T$} on average set size on CIFAR100-LT (the imbalance factor $\gamma$ is set $0.1$) with THR, APS, and RAPS as non-conformity scores.}
    \label{fig:ablation_temperature}
\end{figure}

\paragraph{How does the temperature parameter $\boldsymbol{T}$ affect the performance of conformal training?}
It is important to assess whether making Split CP differentiable for backpropagation through the objective and model parameters impacts model performance with respect to prediction set size and coverage gap. In this paper, we use a sigmoid function to smooth the indicator function in the definition of the prediction set and a temperature parameter $T$ to control the smoothness of this approximation (\eg, as $T \to \infty$, we recover the original formulation).
\Cref{fig:ablation_temperature} ablate the effect of this parameter on CP performance with THR, APS, and RAPS. This experiment is conducted on CIFAR100-LT. During training, we set the mis-coverage level to $\alpha=0.01$, while inference uses $\alpha=0.1$. The results show that \ours consistently achieves stronger performance, especially when the temperature $T$ is small.
This is expected because as $T$ approaches 0 the sigmoid function becomes a soft approximation of the indicator function.
This ablation applies only to \ours, ConfTr~\cite{stutz2022conftr}, InfoCTr~\cite{correia2025infocp}, and DPSM~\cite{shi2025direct}, because these are the only objectives that use a temperature parameter to smooth the indicator during training.

\begin{figure}[!htbp]
    \centering
    \includegraphics[width=\linewidth]{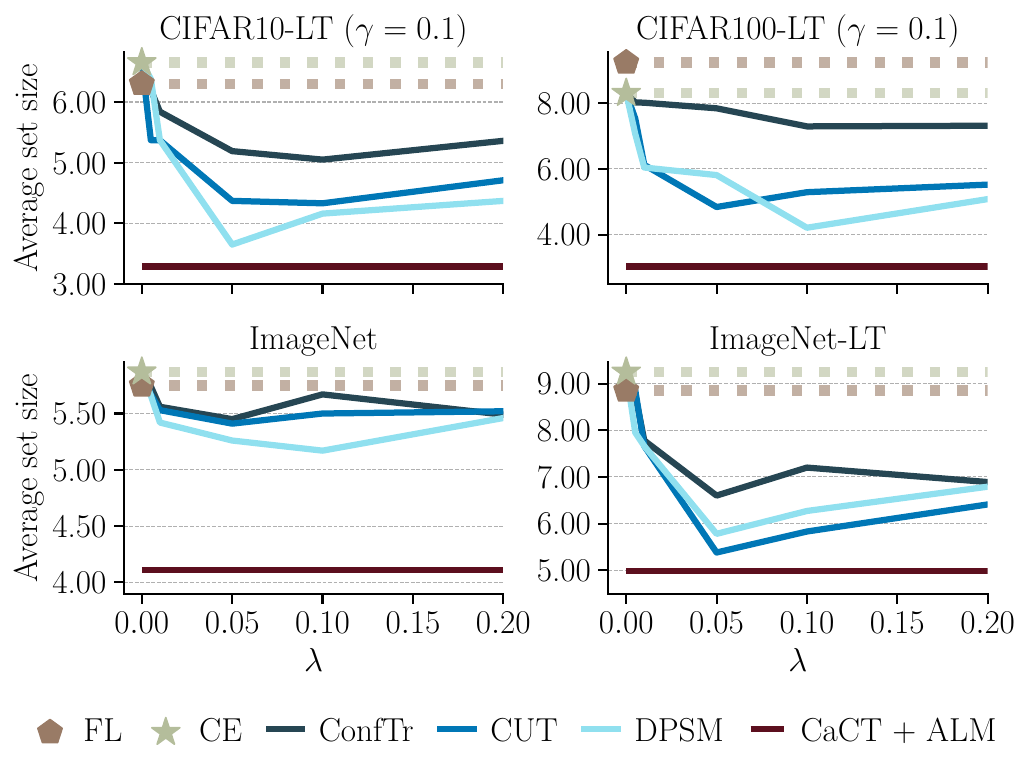}
    \caption{\textbf{Sensitivity to the balancing term $\lambda$ that controls the regularizer.} Selecting an appropriate $\lambda$ for existing methods such as ConfTr~\cite{stutz2022conftr}, CUT~\cite{einbinder2022cut}, and DPSM~\cite{shi2025direct} is challenging, especially on complex datasets like ImageNet-LT. For instance, ConfTr achieves its best performance on ImageNet with $\lambda=0.10$, but requires $\lambda=0.05$ on ImageNet-LT. In contrast, our proposed Augmented Lagrangian formulation eliminates the sensitivity to this hyperparameter, as it is learned adaptively during training, while simultaneously delivering superior results.} 
    \label{fig:multipliers_v2}
\end{figure}

\paragraph{Sensitivity to the regularization weight $\lambda$.}
\Cref{fig:multipliers_v2} evaluates the effect of different regularization weights, fixed during training, on existing approaches when evaluated on long-tailed datasets. These results show that even small changes in~$\lambda$ can lead to significant shifts in efficiency. For example, on ImageNet-LT the average set sizes can shift up to two classes, underscoring the instability and non-trivial nature of the tuning process. \ours is therefore a natural choice for large scale and class-imbalanced settings because it eliminates the need to select a regularization weight in advance. Through ALM, \ours learns to adjust the weights adaptively and in a manner that reflects the difficulty of each class.


\end{document}